%% file: main.tex
\documentclass[final]{anthology-ch}         

\usepackage{framed}
\usepackage{xcolor}
\definecolor{burgundy}{RGB}{128,0,32}
\usepackage{booktabs}
\usepackage{float}
\usepackage{graphicx}
\usepackage{multirow}
\usepackage{makecell}
\usepackage[numbers,sort&compress]{natbib}

\usepackage{subcaption} 
\usepackage{tikz}       
\usepackage{siunitx}    

\newcommand{\panel}[3][0.48\linewidth]{%
  \begin{tikzpicture}
    \node[inner sep=0] (img) {\includegraphics[width=#1]{#2}};
    \node[anchor=north west, xshift=-15pt, yshift=-6pt, fill=white, rounded corners=2pt, inner sep=2pt, text=black, font=\bfseries] 
         at (img.north west) {#3};
  \end{tikzpicture}%
}

\title{Foundation Models Knowledge Distillation For Battery Capacity Degradation Forecast}

\author[1]{Joey Chan}[
  orcid=0009-0004-8100-1546
]

\author[1]{Zhen Chen}[
  orcid=0000-0003-2590-0307
]

\author[1]{Ershun Pan}[
  orcid=0000-0001-6026-9755
]

\affiliation{1}{
  Department of Industrial Engineering and Management, School of Mechanical Engineering, Shanghai Jiao Tong University, Shanghai 200240, China
}


\pubyear{2025}
\pubvolume{1}
\pagestart{1}
\pageend{1}
\conferencename{Proceedings of Conference XXX}
\conferenceeditors{Editor1 Editor2}
\doi{00000/00000}  

\begin{document}

\maketitle

\input{Sections/Abstract}

    \input{Sections/Introduction}
    \input{Sections/Dataset}

    \input{Sections/Methods}

    \input{Sections/Experiments}
    \input{Sections/Discussion}

    \input{Sections/Conclusion}
    \input{Sections/materials}
    \input{Sections/Acknowledgment}

\bibliographystyle{unsrt}
\bibliography{main.bib}


\end{document}

%% file: Sections/Abstract.tex
\begin{abstract}
    Accurate forecasting of lithium-ion battery capacity degradation is critical for reliable and safe operation, yet remains challenging under distribution shifts across scales and operating regimes. 
    We investigate a time-series foundation model—a large pre-trained timeseries model for capacity degradation forecasting, and propose a degradation-aware fine-tuning strategy that aligns the model to capacity trajectories while retaining broadly transferable temporal structure. 
    We instantiate this approach by fine-tuning the Timer model on 220,153 cycles of open-source charge–discharge records to obtain Battery-Timer. 
    Using our released CycleLife-SJTUIE dataset, a real-world industrial collection from an energy-storage station with long-horizon cycling, we evaluate capacity generalization from small cells to large-scale storage systems and across varying conditions. 
    Battery-Timer consistently outperforms specialized expert models; 
    To address deployment cost, we further introduce knowledge distillation, a teacher–student transfer that compresses the foundation model’s behavior into compact expert models. 
    Distillation across several state-of-the-art time-series experts improves multi-condition capacity generalization while substantially reducing computational overhead, indicating a practical path to deployable cross-scale degradation forecasting by combining a foundation model with targeted distillation.
\end{abstract}



%% file: Sections/Introduction.tex
\section{Introduction}

Accurate forecasting of battery capacity is fundamental to ensuring the safety, efficiency, and longevity of lithium-ion batteries (LIBs).
As cycling proceeds, irreversible degradation accumulates toward failure\cite{wang2023explainability}.
In battery In battery PHM, state-of-health (SOH) and remaining useful life (RUL) are two facets of the same degradation process and can be operationalized via the evolution of capacity, making capacity forecasting a natural unifying target.
forecasting the degradation trajectory not only facilitates proactive maintenance and optimal energy management but also informs critical decisions in battery design, reuse strategies, and second-life applications\cite{wang2024physical}.
Therefore, research on battery capacity forecasting is not only technically essential but also strategically important for accelerating the adoption of intelligent and reliable energy storage technologies\cite{chen2024hybrid}.

In recent years, a variety of approaches have been proposed for capacity forecasting, driving significant progress in the field of battery health management.
These approaches are typically classified into two major categories: model-based, data-driven and hybrid methods \cite{guo2022review}.
Model-based techniques rely on empirical formulations or physics-based equations to characterize the capacity degradation process. Although these methods offer valuable interpretability and insights into the underlying degradation mechanisms, they often fail to capture the complex and nonlinear dynamics inherent in long-term degradation sequences.
Moreover, the extraction and selection of degradation-related features remain labor-intensive and require substantial domain expertise \cite{meng2019review}.
To overcome these limitations, data-driven methods have attracted increasing attention. 
Classical machine learning algorithms—such as Support Vector Machines (SVM) \cite{burges1998tutorial}, Random Forests, and Gradient Boosting Machines (e.g., XGBoost\cite{chen2016xgboost})—leverage structured input features through feature engineering and are particularly suitable for scenarios involving small sample sizes or high-dimensional data.
Physics-informed hybrid time-series forecasting model extracts the most informative signals from capacity-degradation sequences and learns them with a data-driven model\cite{chen2024novel}.
While hybrid data-driven models have made strides by combining physics-informed signals and machine learning techniques, they still face major limitations in generalizing across the full diversity of battery configurations and real-world operational modes.
Most existing approaches are tuned to specific datasets or cell types, resulting in performance drop-offs when applied to previously unseen batteries, chemistries, or temperature ranges\cite{cai2024dual,li2023co}.
Despite existing condition-specific branch strategies \cite{che2022data} and domain-invariant feature extraction techniques \cite{guo2025uncovering} demonstrated their effectiveness.
Branching and feature extraction introduce additional engineering overhead and fall short of enabling a simple, reproducible deployment path.

With the rapid progress of Transformer-based\cite{vaswani2017attention} sequence modeling technologies, researchers have explored two related but distinct lines: directly repurposing large language models (LLMs) for non-text signals, and developing time-series foundation models (TSFMs). 
While LLMs operate on discrete tokens for text generation, TSFMs adopt attention and autoregressive next-step learning for numeric sequences, enabling multi-horizon forecasting rather than language modeling. Recent studies illustrate both directions: some reprogram LLM input and output to treat signals as tokenized blocks for inference, for example Pang \textit{et al.} \cite{pang2024hybrid}; 
others design and fine-tune TSFMs for temporal tasks, such as Chronos with parameter-efficient fine-tuning \cite{gupta2024beyond} and unified time-series tasks with pretrained models \cite{zhou2023one}. 
Notable TSFMs such as TimeGPT\cite{garza2023timegpt}, Timer\cite{liu2024timer}, and Lag-Llama\cite{rasul2023lag} learn general-purpose temporal priors from large, heterogeneous corpora and support zero-shot (ZSL) transfer across downstream tasks. 
This paradigm reduces task-specific development and provides broad temporal structure—long-range dependencies, trend and seasonality, and horizon-aware error control—that can be adapted through lightweight fine-tuning.

Although fine-tuned large models show strong adaptability and can be parallelized to handle domain-specific tasks efficiently, their use in LiB health management remains limited. 
Despite the growing attention to TSFMs and their success in learning transferable sequence priors, integration into battery prognostics and diagnostics is still at an early stage. 
A key difficulty is that battery time series differ markedly from conventional forecasting data. 
The timestamp of capacity is observed on a cycle index rather than a uniform wall-clock grid. 
Heterogeneity across batteries further compounds the challenge. 
Chemistries such as LFP, NMC, and NCA, different form factors and electrode architectures, capacity classes from small cells to large energy-storage cells, temperature and C-rate schedules, and manufacturer and BMS differences all reshape the trajectory statistics—slopes, inflection points, and variance growth—causing persistent distribution shift across datasets. 
These properties make cross–capacity generalization in capacity-degradation forecasting substantially more difficult than standard time-series transfer.
At the same time, the substantial parameter sizes of foundation models pose practical barriers to deployment and system integration. 
Their resource-intensive nature limits adoption in edge devices and embedded battery-management systems, where computational efficiency, latency, and power consumption are critical.

In response to these challenges, our goal is to achieve cross–capacity-scale generalization in capacity-degradation forecasting, transferring what is learned on small cells to large energy-storage cells. 
We define zero-shot learning for capacity-degradation forecasting as forecasting under unseen operating conditions without task-specific fine-tuning, for example transferring across capacity scales or charge/discharge protocols.
We proceed in two steps. 
First, we fine-tune a time-series foundation model on public small-cell datasets to obtain a strong teacher with broad temporal priors. 
Second, to satisfy onboard constraints, we distill this knowledge into compact expert forecasters so that they retain the teacher’s cross-capacity robustness at a fraction of the computational and memory cost.
Following Hinton’s teacher–student paradigm \cite{hinton2015distilling}, the expert model is trained with a hybrid objective that combines hard labels with soft targets produced by the pre-trained time-series foundation model. 
This strategy guides the expert to reproduce the teacher’s long-horizon behavior, improves generalization under cross-capacity distribution shift, and supports limited zero-shot transfer when labeled data are scarce, while maintaining the latency and energy profile required for embedded BMS.
The main contributions of this study are as follows:

\begin{enumerate}
  \item We fine-tune a time-series foundation model using Low-Rank Adaptation (LoRA) for univariate, long-horizon capacity forecasting and evaluate transfer from small cells in the pretraining phase to large energy-storage cells in the deployment phase. 
  To the best of our knowledge, this is the first study to treat cross-capacity degradation generalization through the lens of a TSFM.
  \item We propose a knowledge-distillation framework that transfers the TSFM’s generalization ability into lightweight expert models suitable for embedded and edge deployment, improving robustness in few-shot and cross-condition scenarios while respecting onboard resource constraints.
  \item Distillation across multiple expert forecasters shows that a substantial portion of the TSFM’s cross-capacity gains is preserved at much lower inference cost. 
  We release the fine-tuned model and experimental artifacts to facilitate reproducibility and industrial adoption.
\end{enumerate}

\begin{figure}[  htb]
  \centering
  \includegraphics[width=\linewidth]{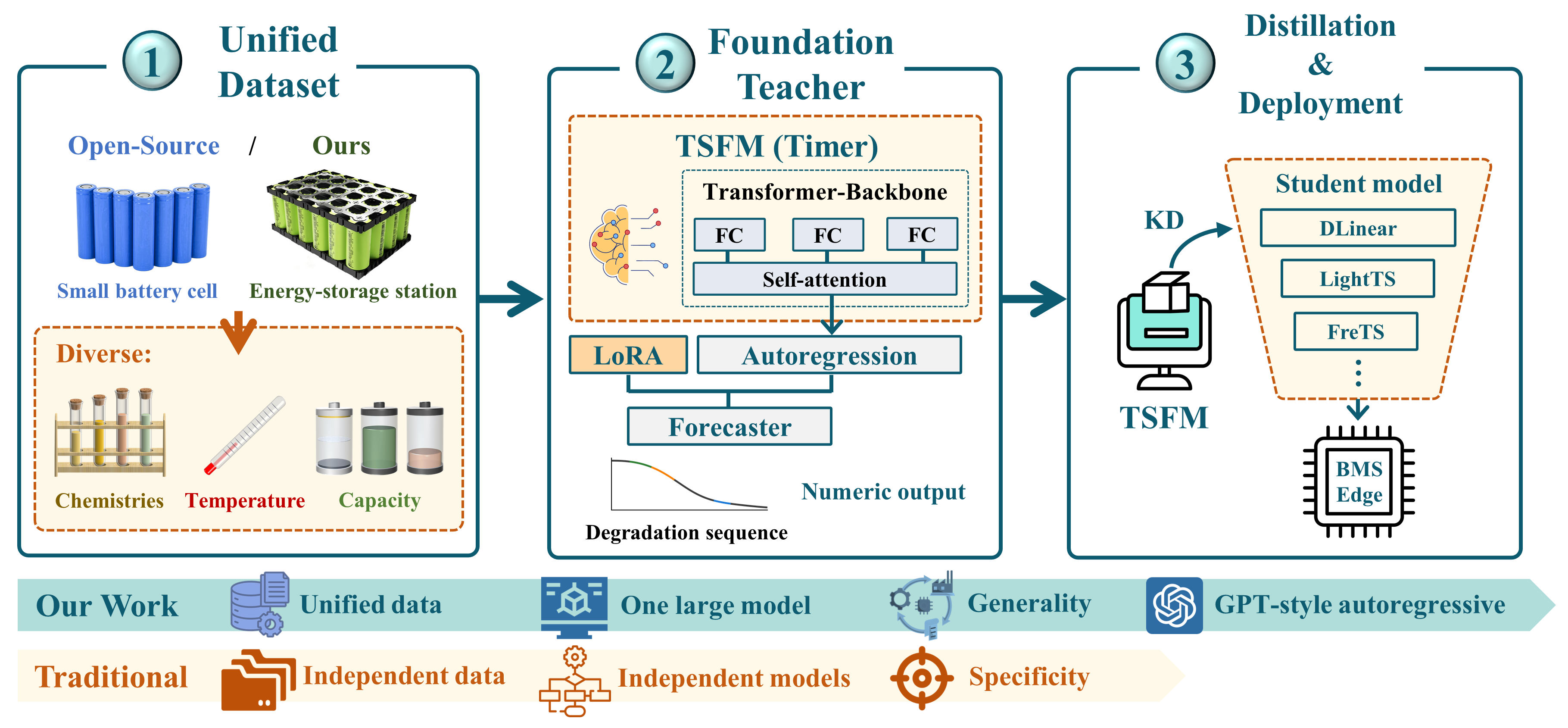}
\caption{Process flow of our study. 
A TSFM is fine–tuned with LoRA on small–cell, cycle–indexed capacity sequences, evaluated in zero–shot on large energy–storage cells and different charging protocols, and distilled into compact expert models for onboard BMS. }
  \label{fig:overview}
\end{figure}

An schematic of the working process is provided in Fig.~\ref{fig:overview}.

This article is organized as follows.
Section 2 gives the problem definition and dataset description.
Section 3 introduces the proposed method in detail.
In Section 4 and 5, the effectiveness of the proposed method is verified.
Section 6 discusses the model forgetfulness and deployment parameter.
Section 7 summarizes this article.

%% file: Sections/Dataset.tex
\section{Preliminary}

\subsection{Data Description}

\begin{figure*}[  htb]
        \center\renewcommand{\figurename}{Figure}
        \includegraphics[width=\textwidth]{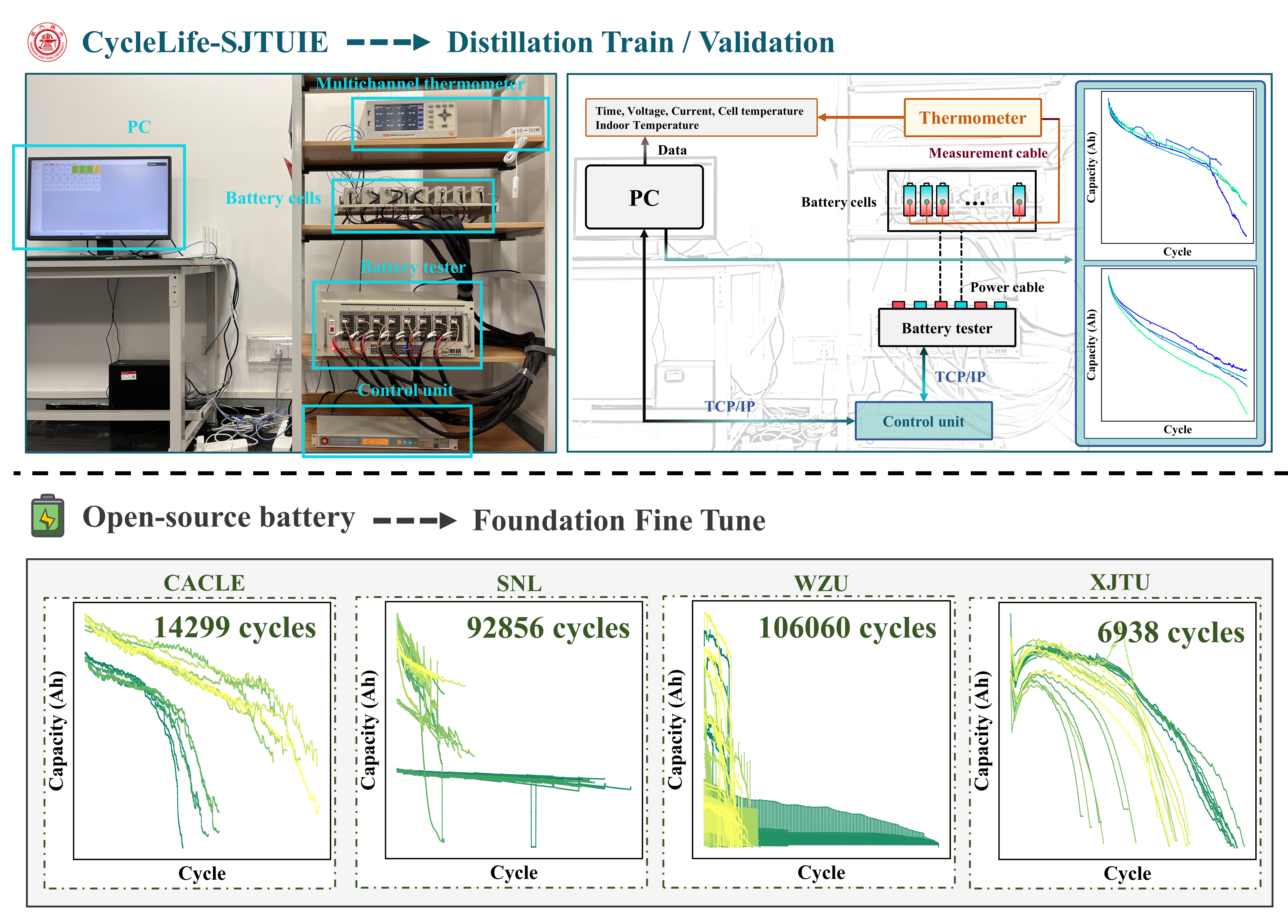}
        \caption{An overview of the datasets used in this study is provided.
            To evaluate the model's generalizability, experiments will be conducted on various battery models.}
        \label{img1}
    \end{figure*}

Large-scale datasets form the foundation for fine-tuning foundation models.
This study divides the dataset into two parts: an open-source battery capacity degradation dataset used for fine-tuning the foundation model, and the SJTUIE-Battery CycleLife dataset from Shanghai Jiao Tong University, which is used for distillation training and testing.
Detailed information about the datasets is provided in Figure \ref{img1}.

\subsubsection{Cyclelife-SJTUIE}

\begin{table}[htb]\caption{Details of the datasets used for performance evaluation.}
    \centering \label{dataset_discription}
    \begin{tabular}{cccc}
        \toprule[1.2pt]
        Attribute       & Value     & Parameter   & Value \\ \hline
        Product name          & IFR32135         & Rated capacity  & 13Ah \\
        Shape          & Cylindricity         & Normal voltage     & 3.2V  \\
        Cathode          & LiFePO4           & Upper cut-off voltage &  3.65V(CCCV), 3.9V (CC)  \\
        Anode          & Graphite           & Lower cut-off voltage    &2V  \\
        \toprule[1.2pt]
\end{tabular}
\end{table}

The CycleLife-SJTUIE dataset was collected through experiments conducted on eight battery cells produced by Gotion High Tech Co., Ltd.
The relevant information about these cells is listed in Table \ref{dataset_discription}.
The cells were placed indoors without temperature control and mounted in Neware A708-4B-J-30A battery fixtures, with each cell connected to one channel of the Neware CTE-4008D-5V30A battery tester for charging and discharging.
The tester recorded current, voltage, and other electrical data, while the control unit, Neware CT-ZWJ-4 ST-1U, managed data collection and tester commands.
Temperature measurements were taken using thermocouples attached to each cell's surface and recorded by a Jinko JK5000-24 temperature tester, with ambient temperature monitored by a Huahanwei TH42W-EX thermometer.
Cells \#1–4 underwent constant current (CC) charging, and cells \#5–8 were charged using a constant current and constant voltage (CCCV) method.
A detailed test profile is provided in Profile \ref{operator}.
    
    \begin{table*}[htb]\caption{A test profile for CC and CCCV.}
    \centering \label{operator}\scriptsize
    \begin{tabular}{l}
        \toprule[1.2pt]
        1: \textbf{Initialization:} Discharge all the cells completely.  \\
        2: \textbf{while} current capacity/rated capacity > 0.8 \textbf{do} \\
        3: (CC profile) CC charge at 1C (13A) until the voltage reaches 3.9V;\\
        (CCCV profile) CC charge at 1C (13A) until the voltage reaches 3.65V, and then CV charge at 3.65V until the current drops to 0.05C (0.65A).  \\
        4: Rest 30 min.  \\
        5: CC discharge at 1C (13A) until the voltage drops to 2V.  \\
        6: Rest 30 min.\\
        7: end while\\
        \toprule[1.2pt]
    \end{tabular}
    \end{table*}

\subsubsection{Open-source battery capacity degradation dataset}

This study collects and utilizes four publicly available battery datasets, as illustrated in Table \ref{tab:selected_datasets}.
From approximately 220,153 cycles of raw charge–discharge records, we extract capacity degradation sequences, which are subsequently used for fine-tuning the foundation model described in Figure\ref{img2}.

\input{Sections/Tables/data_table.tex}

\subsection{Foundation model: Timer}

\begin{figure}[htb]
    \center\renewcommand{\figurename}{Figure}
    \includegraphics[width=0.6\textwidth]{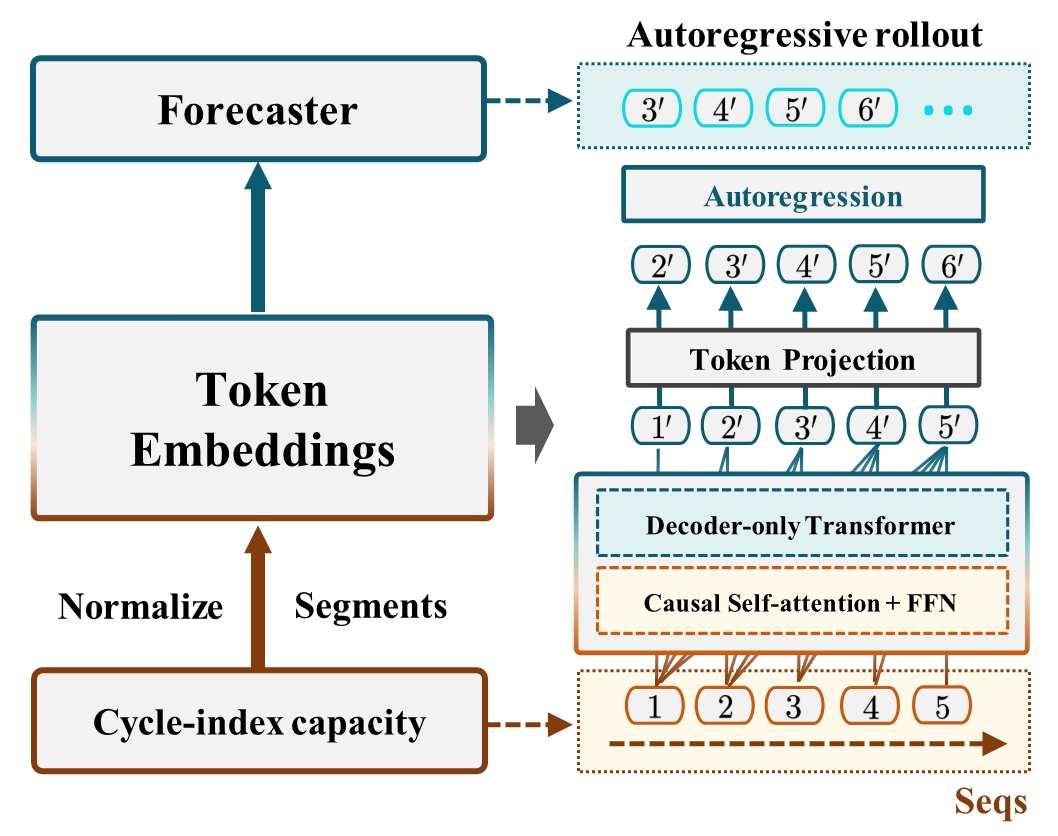}
    \caption{Architectures of Timer: a Transformer-decoder-based forecasters.}
    \label{img2}
\end{figure}

The time-series foundation model used in this study is \textbf{Timer}\cite{liu2024timer}, a Transformer\cite{vaswani2017attention}-based architecture developed by Tsinghua University.
To build a scalable backbone for large time series models (LTSMs),Timer adopts a decoder-only Transformer architecture in Figure\ref{img2}, inspired by the autoregressive generation paradigm used in LLMs.
Given a time series sequence $X = \{x_1, x_2, ..., x_{NS}\}$, Timer tokenizes it into non-overlapping segments of length $S$ as \textbf{Eq.}\ref{eq1}.
\begin{equation}\label{eq1}
    s_i = \{x_{(i-1)S+1}, ..., x_{iS}\} \in \mathbb{R}^S, \quad i = 1, ..., N,
\end{equation}
where $NS$ is the total sequence length and $N = NS / S$ is the number of tokens.
In \textbf{Eq.}\ref{eq2} the token embeddings are passed through $L$ layers of Transformer blocks with causal masking.
\begin{align}\label{eq2}
&h_i^0 = W_e s_i + \text{TE}_i, \\
&h^l = \text{TrmBlock}(h^{l-1}), \quad l = 1, ..., L, \\
&\hat{s}_{i+1} = W_d h_i^L,
\end{align}

where $W_e, W_d \in \mathbb{R}^{D \times S}$ are the input/output projection matrices, $\text{TE}_i$ denotes optional temporal embedding, and $h_i^L$ is the final hidden representation.
The model is trained with the generative objective as \textbf{Eq.}\ref{eq3}.
\begin{equation}\label{eq3}
\mathcal{L}_{\text{MSE}} = \frac{1}{NS} \sum_{i=2}^N \|s_i - \hat{s}_i\|_2^2,
\end{equation}
which enforces token-wise supervision across all predicted tokens.

This design enables Timer to support flexible context lengths, autoregressive inference, and multi-step forecasting with sliding windows.

%% file: Sections/Tables/data_table.tex
\begin{table*}[htbp]
\centering
\scriptsize
\caption{Selected Battery Datasets (WZU, CALCE, SJTU, XJTU)}
\label{tab:selected_datasets}
\begingroup
\sisetup{
  detect-all,
  group-separator = {,},
  group-minimum-digits = 4,
  input-ignore = {,},
  table-number-alignment = center
}
\setlength{\tabcolsep}{3pt}
\renewcommand{\arraystretch}{1.05}


\begin{tabular}{
  @{}l  l  l
  S[table-format=3.0]
  S[table-format=7.0]
  l  l  l@{}
}
\Xhline{1.2pt}
\textbf{Dataset} & \textbf{Cathode} & \textbf{Anode} &
\textbf{Cell Num} & \textbf{Cycle Num} & \textbf{Capacity (Ah)} &
\textbf{Temperature (\si{\celsius})} & \textbf{Format} \\
\hline
WZU\cite{wang2023large}   & LFP      & Graphite & 242 & 106,060 & 1 / 0.8 / 3 & Seasonal   & Cells + parallel packs \\
CALCE\cite{xing2013ensemble} & LCO      & Graphite & 13  & 14,299  & 1.10         & 25         & Prismatic \\
XJTU\cite{wang2024physics}  & NMC532   & Graphite & 23  & 6,938   & 2.00         & 20         & 18650 cyl. \\
SJTU  & LFP/NMC  & Graphite & 8   & 487,200 & 13.0         & Room Temp  & IFR32135 \\
\Xhline{1.2pt}
\end{tabular}
\endgroup 

\end{table*}

%% file: Sections/Methods.tex
\section{Our Method: Degradation Knowledge Distillation from Foundation Models}

\begin{figure}[H]
  \centering
  \panel[0.85\linewidth]{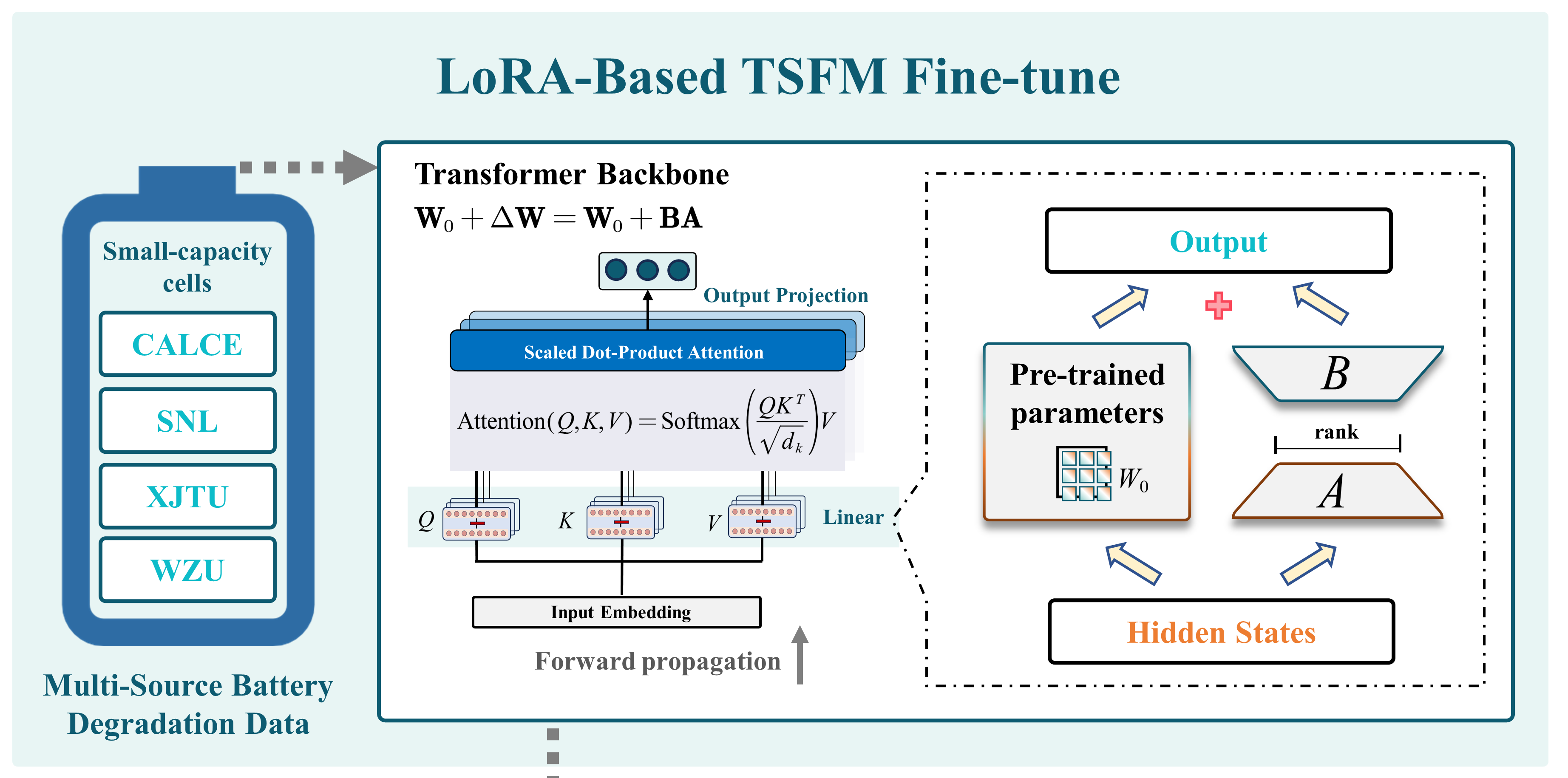}{(a)}\\
  \panel[0.85\linewidth]{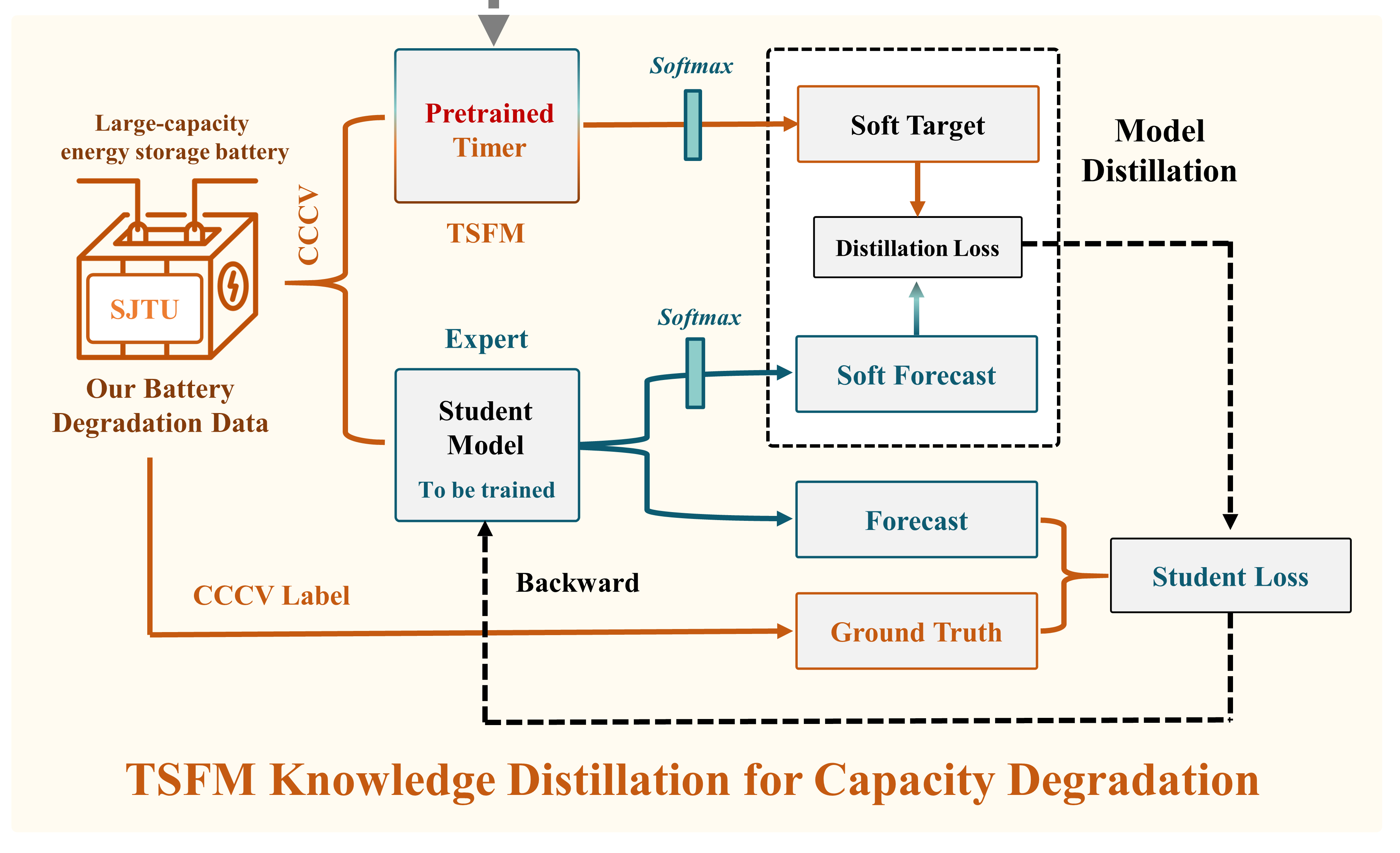}{(b)}
  \caption{\textbf{Our Method.} 
  \textbf{(a)} LoRA fine-tuning and cross-protocol deployment. 
    A time-series foundation model is fine-tuned with Low-Rank Adaptation on degradation data from small-capacity cells to enable transfer to large-capacity energy-storage cells. 
  \textbf{(b)} Response-Based Knowledge Distillation Framework for Time-Series Foundation Models.
        An expert model distilled under the CCCV protocol is then applied to CC operation to evaluate cross-protocol generalization.}
  \label{img3}
\end{figure}

\subsection{Battery Degradation Fine-Tuning}

Fine-tuning aligns the generic temporal priors of a pre-trained time-series foundation model with the cycle-indexed capacity–degradation forecasting task, achieving a balance between prior knowledge and battery-specific dynamics.
Currently, Low-Rank Adaptation (LoRA) has emerged as one of the most popular fine-tuning techniques due to its significant reduction in memory usage and computational cost \cite{hu2022lora}.

In this study, we adapt the foundation model (\emph{Timer}) to discharge–capacity degradation forecasting and apply LoRA to the Timer model by injecting trainable parameters into the modules:
\textcolor{burgundy}{\texttt{"q\_proj"}, \texttt{"v\_proj"}, \texttt{"k\_proj"}}, as shown in Figure \ref{img3} (a). 
For the 84M-parameter Timer model released by Tsinghua University, the learnable LoRA parameters account for only 0.4673\% of the total.
LoRA introduces two low-rank matrices $\mathbf{A}\!\in\!\mathbb{R}^{d\times r}$ and $\mathbf{B}\!\in\!\mathbb{R}^{r\times d}$ that form a low-rank update $\Delta\mathbf{W}=\mathbf{B}\mathbf{A}$; the backbone weight is then used as $\mathbf{W}_0+\Delta\mathbf{W}$ so that the base $\mathbf{W}_0$ remains frozen while only $\mathbf{A},\mathbf{B}$ are trained.

\begin{equation}\label{equ4}
\mathbf{W}_0+\Delta \mathbf{W}=\mathbf{W}_0+\mathbf{BA}
\end{equation}

To encode the physical tendency of capacity to decrease over cycles, we augment the objective with a trend penalty that discourages upward drifts in the predicted sequence.
In addition to the standard mean squared error loss $\mathcal{L} _{\mathrm{MSE}}=\frac{1}{N}\sum_{i=1}^N{\left( y_i-\hat{y}_i \right) ^2}$, we add a Sigmoid-based penalty on the end-to-start difference to penalize sequences exhibiting upward trends, thereby promoting the generation of degrading sequences. 
This acts as a lightweight, physics-informed soft constraint during fine-tuning.

\begin{equation}\label{equ5}
\begin{aligned}
\mathcal{L} &=\mathcal{L}_{\mathrm{MSE}}+\lambda \mathcal{L}_{\mathrm{trend}}
\\
\mathcal{L}_{\mathrm{trend}}&=\mathrm{Sigmoid}\!\left( \hat{y}_N-\hat{y}_1 \right) 
\end{aligned}
\end{equation}

At the end of this paper, we release the trained LoRA parameters for battery degradation modeling.
The provided checkpoints and configuration are compatible with Timer under a 96-step input horizon on cycle-indexed capacity sequences.
We further report an ablation study analyzing how the LoRA injection positions (e.g., \texttt{q\_proj}, \texttt{k\_proj}, \texttt{v\_proj}) affect forecasting performance.

\subsection{Timeseries Foundation Models Knowledge Distillation}

Recent TSFMs demonstrate strong long-horizon forecasting performance owing to their scalability and capacity to encode long-range dependencies \cite{gou2021knowledge}. 
However, deploying such models in resource-constrained battery-management systems (BMS)—including vehicular and station-level controllers—remains challenging due to latency, memory, and power limits.
Knowledge distillation (KD) offers an effective approach to transferring knowledge from large-scale teacher models to lightweight student models.
Although TSFMs are rapidly advancing, few studies distill knowledge from a time-series foundation model into compact expert forecasters for battery capacity degradation, particularly under cycle-indexed, single-series inputs.
The time-series knowledge distillation framework proposed in this study is illustrated in Figure \ref{img3} (b).

Specifically, we adopt a response-based teacher–student KD scheme tailored to capacity–degradation forecasting: the Battery-Timer TSFM (teacher) produces multi-horizon predictions from a univariate, cycle-indexed input window, and the student is trained to mimic these outputs while fitting ground truth.
The student model $f_{\mathrm{student}}$ learns from both the ground-truth labels and the softened outputs generated by the pre-trained teacher model $f_{\mathrm{teacher}}$.
Given $\hat{Y}_{\mathrm{student}}=f_{\mathrm{student}}\!\left( X \right)$ and $\hat{Y}_{\mathrm{teacher}}=f_{\mathrm{teacher}}\!\left( X \right)$, the student loss and distillation loss are computed as shown in \textbf{Eq.}\ref{equ6}, where $T$ is the temperature used to produce softened targets via $\mathrm{Soft}\max(\cdot/T)$ along the forecast horizon.

\begin{equation}\label{equ6}
    \begin{aligned}
    \mathcal{L}_{\mathrm{hard}}&=\mathcal{L}_{\mathrm{MSE}}\!\left( \hat{Y}_{\mathrm{student}},\,Y \right),\\
    \mathcal{L}_{\mathrm{soft}}&=T^2\cdot \mathrm{KL}\!\left( \mathrm{Soft}\max \!\left( \frac{\hat{Y}_{\mathrm{teacher}}}{T} \right) \parallel \mathrm{Soft}\max \!\left( \frac{\hat{Y}_{\mathrm{student}}}{T} \right) \right).
    \end{aligned}
\end{equation}

The final total loss is defined as a weighted sum of the two components in \textbf{Eq.}\ref{equ7}, where $\alpha$ controls the relative contribution of the soft-label and hard-label losses.

\begin{equation}\label{equ7}
\mathcal{L}_{\mathrm{total}}=\alpha \cdot \mathcal{L}_{\mathrm{soft}}+(1-\alpha)\cdot \mathcal{L}_{\mathrm{hard}}.
\end{equation}

In our setting, the expert student preserves the input/output horizons of the foundation teacher (e.g., 96-step window and forecast), operates on single-series, cycle-indexed capacity sequences without exogenous covariates, and targets deployment under BMS resource constraints while benefiting from the teacher’s cross–capacity-scale generalization (from small cells to large energy-storage cells).

%% file: Sections/Experiments.tex
\section{Experiments}

\subsection{Experiment Setup}

In this section, we rigorously evaluate the effectiveness of our TSFM for battery capacity-degradation forecasting and its KD into resource-efficient expert models, and we outline the overall experimental workflow.
First, to validate the TSFM, we train it on the datasets summarized in Table \ref{dataset_discription} and benchmark it against eight state-of-the-art time-series forecasters. 
Notably, each baseline expert is specifically trained on the SJTU dataset, whereas the TSFM has no exposure to SJTU prior to validation, thereby testing zero-shot transfer. 
To further assess generalization under data scarcity, we conduct leave-one-battery-out (LOBO) experiments in which, from the four training corpora, one battery subset is held out in turn for evaluation. 
The second part of our study examines degradation-sequence knowledge distillation: using the framework in Figure \ref{img3} (b), we distill the TSFM under the CCCV protocol and compare distilled experts with their non-distilled counterparts under both CCCV and CC settings. 
The next section presents ablation studies and hyperparameter analyses that further substantiate the effectiveness of the proposed approach.
The experimental parameters used in this study are summarized in Table \ref{tab:training_parameters}.

\input{Sections/Tables/parameter.tex}

\subsection{Validation for Battery Timer}

Before conducting knowledge distillation from the foundation model, we first ensure that the Timer model has acquired a correct understanding of battery degradation prediction.
To this end, we compare the performance of the fine-tuned model and the original Timer model on the CycleLife-SJTUIE dataset.
Our evaluation protocol is as follows. 
Because the CycleLife-SJTUIE dataset was completely unseen during training of Battery-Timer, all eight cells measured under the two charge–discharge protocols (CC and CCCV) are eligible for validation. 
We adopt a window-based procedure: each degradation trajectory is partitioned into contiguous windows of length 192. 
Given that the forecaster was trained with a base horizon of 96, the first 96 time steps in each window serve as the input context and the remaining 96 steps constitute the forecast target. 
The same procedure is applied uniformly to all baseline methods.

\subsubsection{Comparison with state-of-the-art Method}

To evaluate the superiority of our approach, we compare against eight representative baselines spanning three architectural families: 
MLP-based models and variants (DLinear\cite{zeng2023transformers}, FiLM\cite{zhou2022film}, FreTS\cite{yi2023frequency}, PaiFilter\cite{yi2024filternet}, TSMixer\cite{chen2023tsmixer}, LightTS\cite{zhang2207less}), a Transformer-based variant (PatchTST\cite{nie2022time}), and a recurrent model (SegRNN\cite{lin2023segrnn}). 
These methods reflect recent advances in time-series forecasting. 
All baselines are trained in a supervised manner on the CycleLife-SJTUIE dataset, using the first 60\% of each cell’s trajectory for training and the remaining 40\% for evaluation, following the same validation protocol as the TSFM.

\begin{figure}[  htb]
  \centering
  \panel[0.45\linewidth]{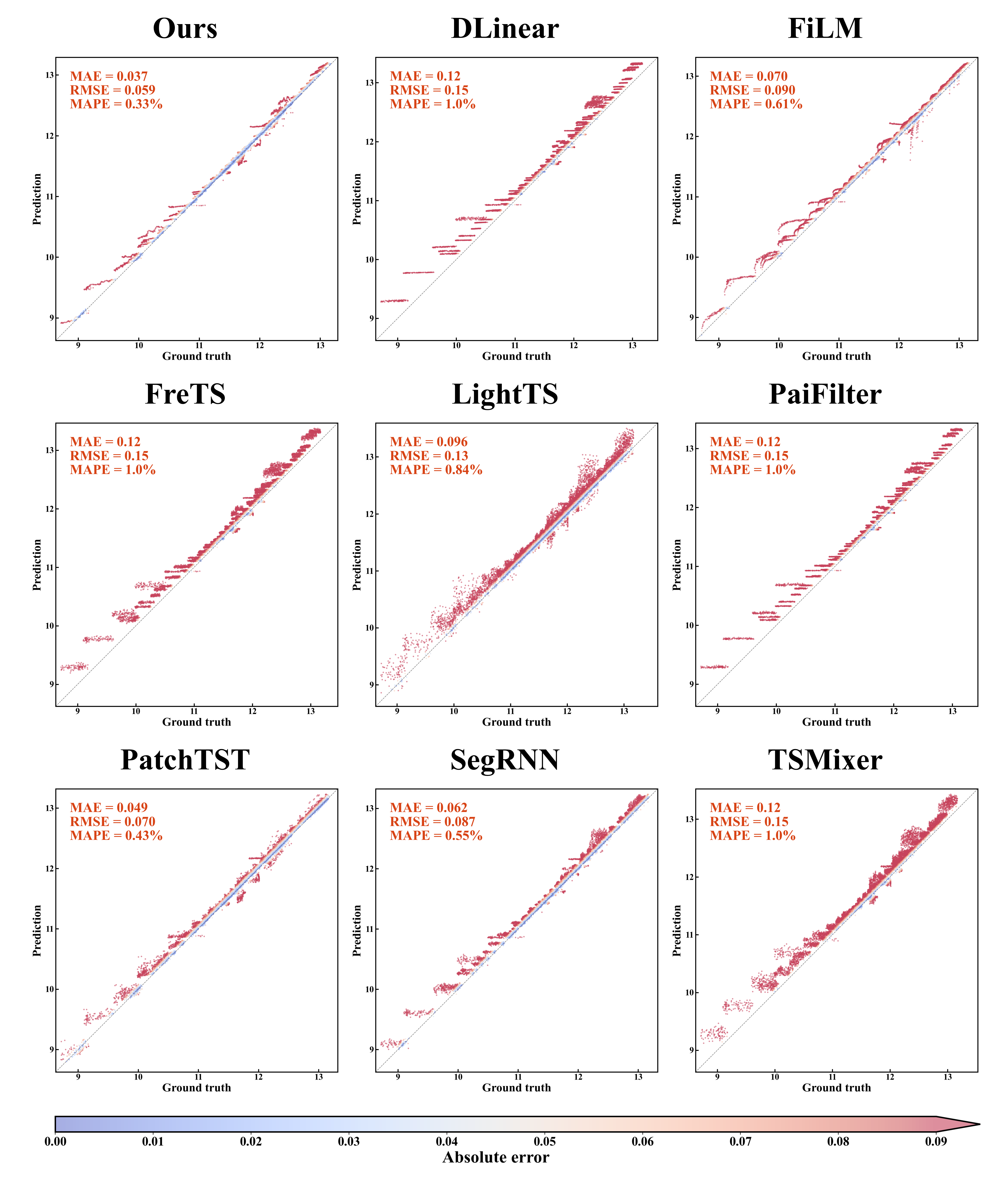}{(a)}
  \panel[0.45\linewidth]{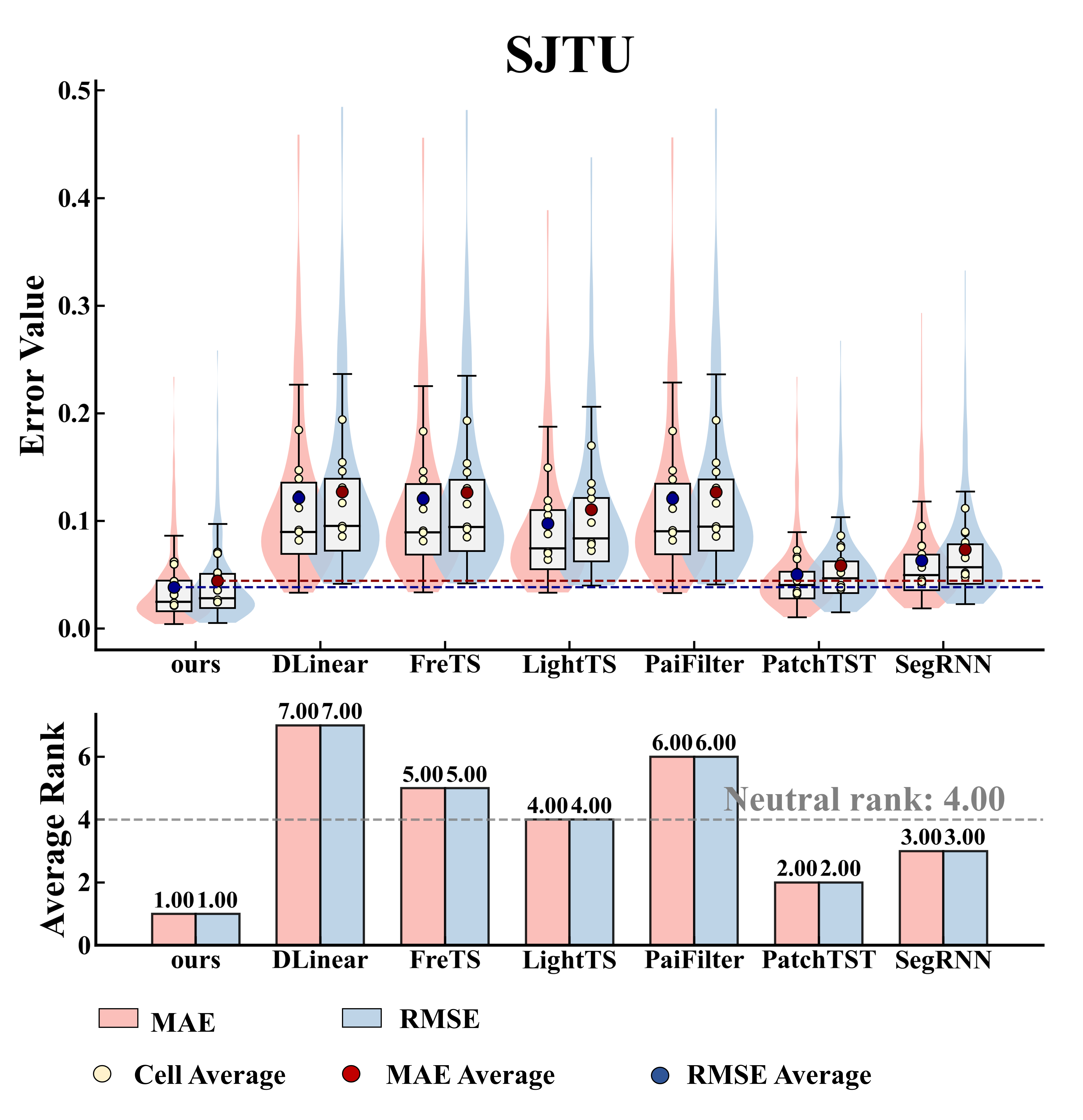}{(b)}
  \caption{\textbf{Performance comparison between Battery-Timer and recent time-series forecasters.} 
  \textbf{(a)} Scatter of predictions versus ground truth under window-based validation; 
  points from our method cluster near the identity line (y = x), indicating higher accuracy.
  \textbf{(b)} Violin plots of average errors across eight cells with corresponding Friedman ranks; 
  the TSFM (Battery-Timer), trained on large-scale data, demonstrates a consistent advantage.}
  \label{c_with_state-of-the-art}
\end{figure}

As shown in Fig.~\ref{c_with_state-of-the-art}, our approach achieves the lowest errors in window-based validation. 
In Fig.~\ref{c_with_state-of-the-art}(a), Battery-Timer attains the best mean absolute error (MAE) and mean absolute percentage error (MAPE) among all baselines. Notably, despite having no training exposure to large-capacity energy-storage cells, it delivers competitive zero-shot performance, underscoring the strength of Transformer-based TSFMs for numerical sequence forecasting. 
Fig.~\ref{c_with_state-of-the-art}(b) reports the average MAE and root mean squared error (RMSE) across two protocols and eight cells; the error distributions further indicate a clear advantage for our method. 
A Friedman ranking places the TSFM first across methods, providing statistical support for its robustness in this setting.

\subsubsection{LOBO Analsis}

To probe whether the TSFM learns cross-dataset invariants of capacity degradation and to assess robustness under data scarcity, we conduct LOBO ablations on the pretraining/fine-tuning corpus.
Starting from the full setting trained on all four source corpora (“Our Method”), we create four variants, each excluding a single corpus (“w/o …”). The evaluation strictly follows the design in Fig.~\ref{c_with_state-of-the-art}(a).
Comparing the LOBO variants against the full model quantifies the contribution of each corpus and indicates how much of the TSFM’s performance arises from shared degradation structure rather than dataset-specific cues.

\input{Sections/Tables/LOBO.tex}

  As summarized in Table \ref{tab:lobo_results}, removing any single corpus leads to a performance drop for the TSFM, with the most pronounced degradation observed in the w/o WZU setting, where the error increases by approximately 46\%. 
  We attribute this to two factors. First, WZU is the most comprehensive corpus in our training pool and spans the broadest range of operating conditions; 
  its absence substantially narrows the model’s exposure to diverse regimes. Second, the WZU corpus shares battery chemistry (LFP) with CycleLife-SJTUIE, providing stronger degradation commonalities that facilitate transfer. 
  Overall, the LOBO results suggest that, akin to scaling laws observed in language modeling, expanding the breadth and diversity of pretraining data can further improve TSFM performance.

\subsection{Validation for Knowledge Distillation}

To examine whether foundation-model priors can be transferred into lightweight, deployable experts, we distill the TSFM under the CCCV protocol and then evaluate the resulting students under two settings: (i) CCCV (in-protocol) to verify that distillation preserves or improves in-domain accuracy, and (ii) CC (cross-protocol) to test zero-shot transfer. 
Importantly, for the CC condition the vanilla baselines are trained in a fully supervised manner on CC data, whereas the distilled experts have never seen CC during training; 
their CC performance therefore reflects genuine zero-shot cross-protocol generalization. 
All models follow the same cycle-indexed, window-based evaluation used earlier, enabling a controlled comparison across protocols.

\input{Sections/Tables/kd_validation.tex}

Table~\ref{tab:vanilla_vs_distilled_means_multirow} compares distilled students with their vanilla counterparts under CCCV and CC protocols. 
Distillation strengthens most backbones in both settings. Under CCCV, seven of eight models improve concurrently on MAE, RMSE, $R^2$, and MAPE. 
The gains are substantial: PaiFilter improves from MAE 0.094 and RMSE 0.109 to MAE 0.016 and RMSE 0.027, with $R^2$ rising from 0.953 to 0.997 and MAPE falling from 0.77\% to 0.13\%. 
PatchTST advances from MAE 0.036 and RMSE 0.045 to MAE 0.016 and RMSE 0.026, with $R^2$ increasing from 0.992 to 0.997 and MAPE decreasing from 0.30\% to 0.13\%. 
DLinear, FreTS, SegRNN, TSMixer, and FiLM show similar upward trends. 
LightTS is the only counterexample and exhibits modest regressions after distillation on CCCV.

Cross-protocol evaluation on CC further demonstrates transfer. 
Distilled students are trained only on CCCV yet are assessed on CC without any CC exposure, while the vanilla baselines are supervised on CC. 
Even under this zero-shot setting, most students surpass their vanilla counterparts. 
FreTS reduces MAE from 0.148 to 0.057 and RMSE from 0.179 to 0.078, with $R^2$ improving from 0.919 to 0.983 and MAPE declining from 1.34\% to 0.51\%. 
PaiFilter drops MAE from 0.148 to 0.043 and RMSE from 0.179 to 0.067, with $R^2$ increasing from 0.918 to 0.985 and MAPE from 1.34\% to 0.39\%. 
PatchTST and TSMixer also record clear improvements. 
FiLM shows mixed outcomes on CC, with lower MAE and MAPE but slightly higher RMSE and a small reduction in $R^2$. 
LightTS remains the consistent exception, where the distilled variant underperforms the vanilla model.

\begin{figure}[  htb]
  \centering
  \panel[0.45\linewidth]{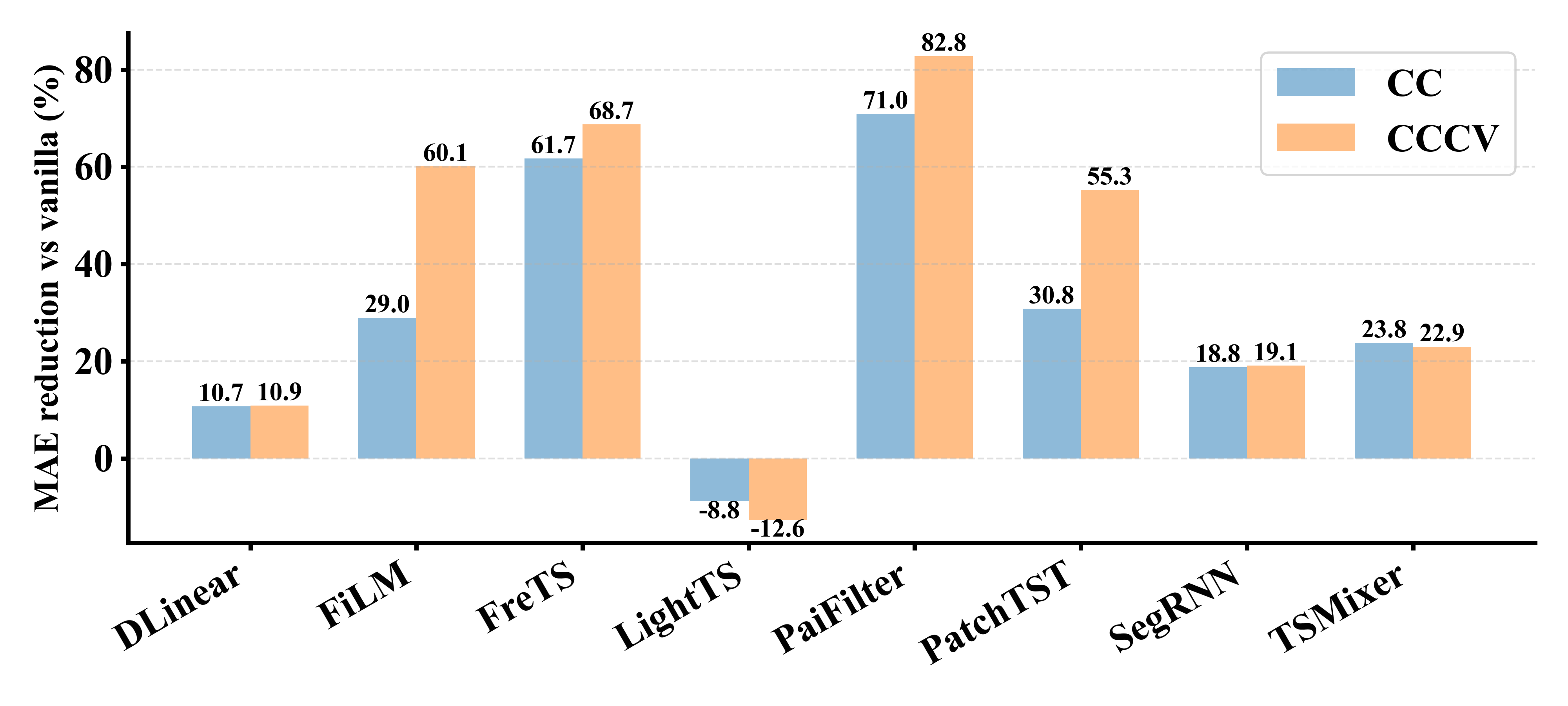}{(a)}
  \panel[0.45\linewidth]{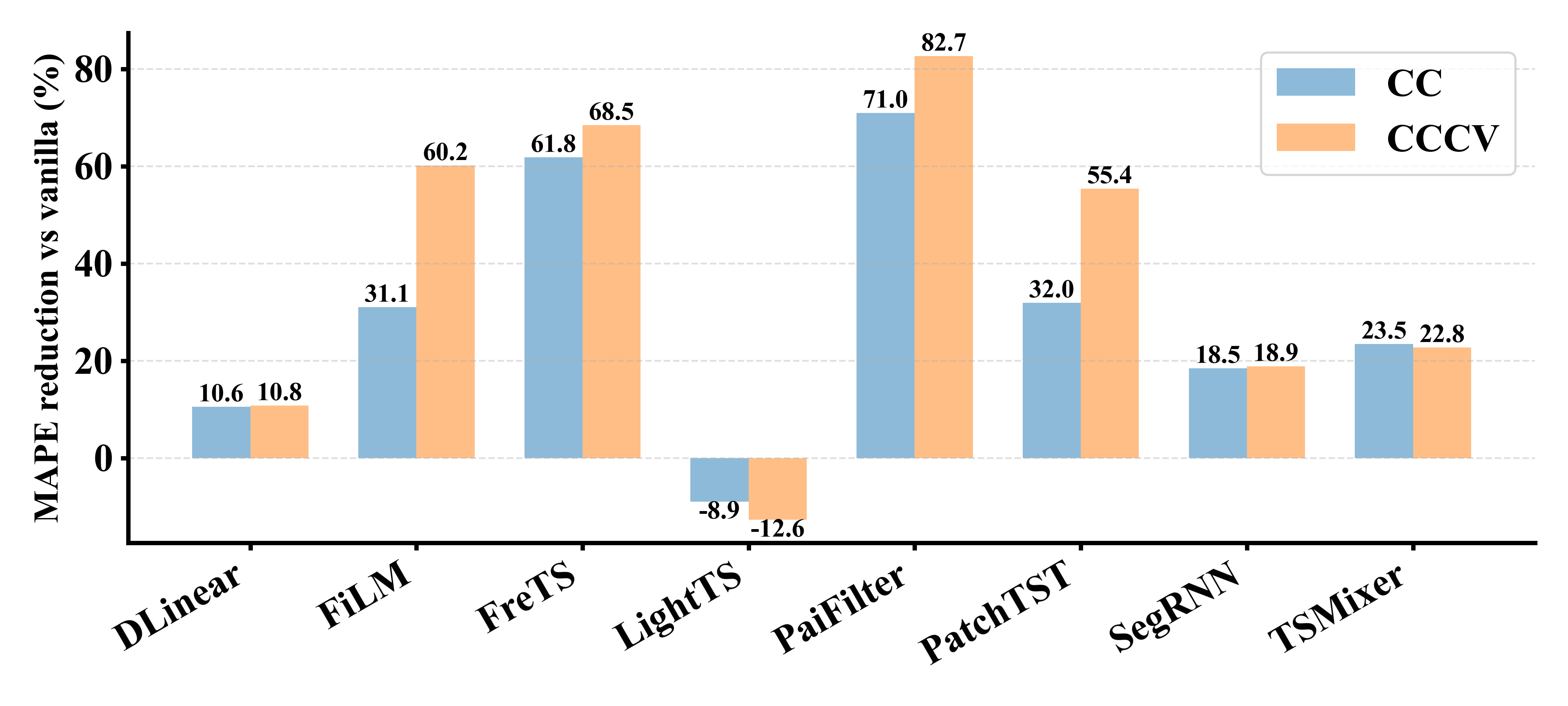}{(b)}\\
  \panel[0.45\linewidth]{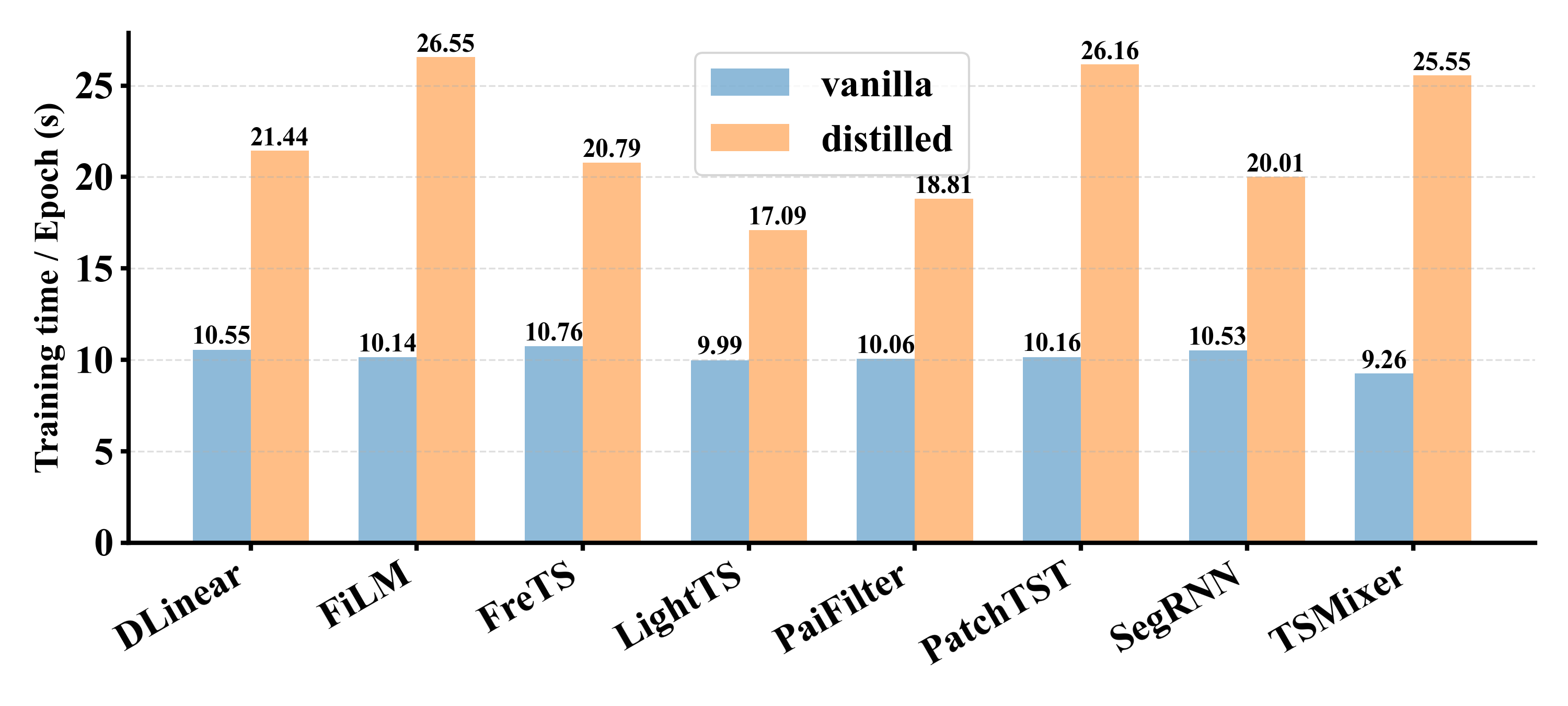}{(c)}
  \panel[0.45\linewidth]{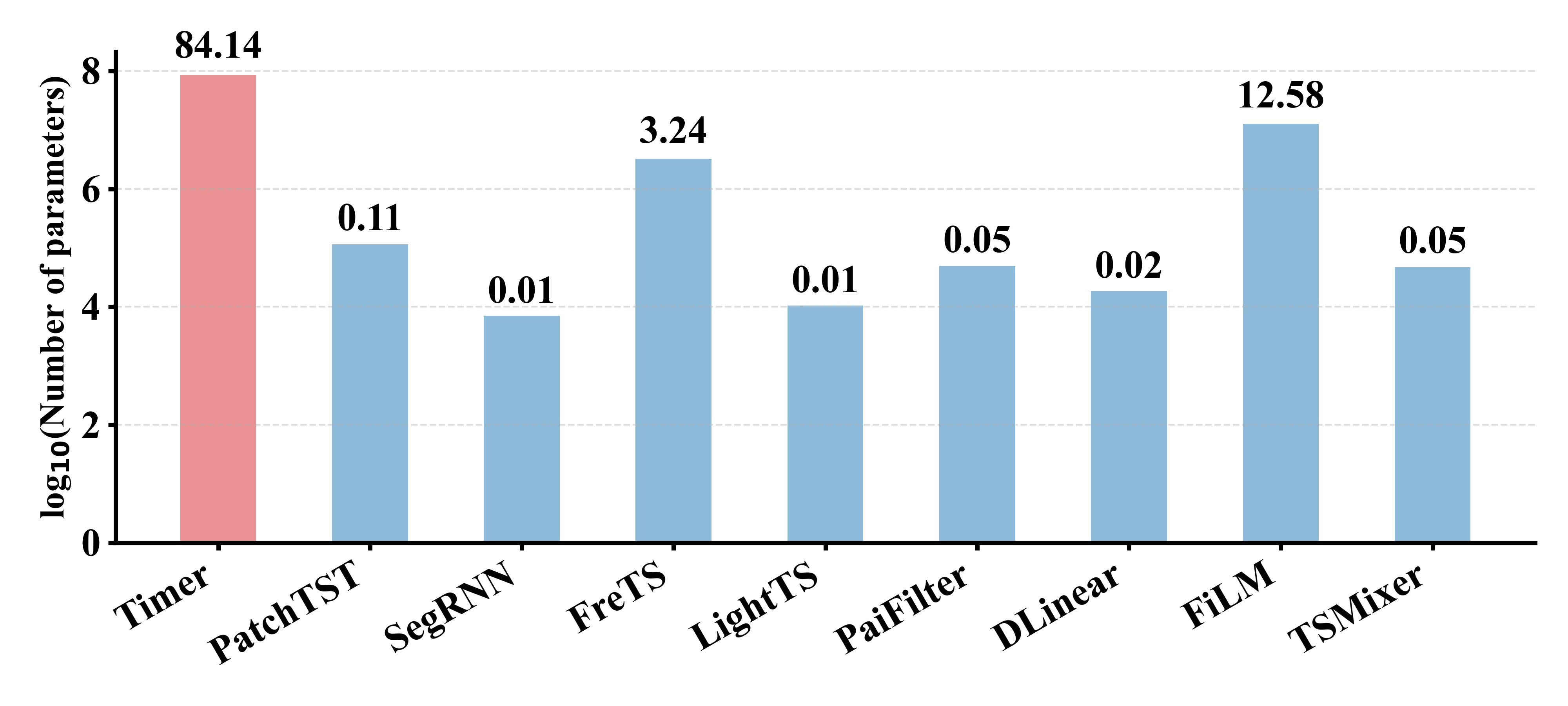}{(d)}
  \caption{\textbf{Changes in computing resources and metrics.} 
  \textbf{(a)} MAE relative change \textbf{(b)} Mape relative change \textbf{(c)} Training time comparison \textbf{(d)} Model parameter sizes}
  \label{metrics_change}
\end{figure}

As shown in Fig.~\ref{metrics_change}, TSFM-based distillation increases wall-clock training time relative to vanilla supervised training, with the average per-epoch time roughly doubling. 
Despite this overhead, models distilled under the CCCV protocol achieve concurrent improvements on both CCCV and CC evaluations and, in several cases, surpass counterparts trained with supervised learning on CCCV. 
In light of these gains, the additional training cost is justified.

Overall, the results indicate that TSFM-to-expert distillation transfers useful temporal priors, improving in-protocol accuracy and enabling zero-shot cross-protocol generalization for most architectures. 
In the Discussion section, we examine the mechanisms by which TSFM-based distillation improves the expert models.

%% file: Sections/Tables/parameter.tex
\begin{table}[htb]
    \centering
    \caption{Experiment Hyperparameters}
    \begin{tabular}{lcc}
    \Xhline{1.2pt}
    \textbf{Parameter} & \textbf{Battery LoRA} & \textbf{KD} \\
    \midrule
    Num epoch          & 5        & 20    \\
    Batch size         & 64       & 4     \\
    Learning rate      & $1 \times 10^{-4}$ & $1 \times 10^{-5}$ \\
    Optimizer          & Adam     & Adam  \\
    LoRA $\alpha$ / $\alpha$ & 16   & 0.3   \\
    Rank               & 8        & None  \\
    LoRA dropout       & 0.05     & None  \\
    \Xhline{1.2pt}
    \end{tabular}
    \label{tab:training_parameters}
\end{table}

%% file: Sections/Tables/LOBO.tex
\begin{table}[htb]
  \centering
  \caption{Leave one battery out results.}
  \label{tab:lobo_results}

  \begin{tabular}{lccccc}
    \toprule
    \textbf{Metric} & \textbf{Our Method} & \textbf{w/o CALCE} & \textbf{w/o SNL} & \textbf{w/o WZU} & \textbf{w/o XJTU} \\
    \midrule
    MAE   & \textbf{0.037} & 0.433 & 0.457 & 0.040 & 0.054 \\
    RMSE  & \textbf{0.059} & 0.067 & 0.082 & 0.065 & 0.091 \\
    MAPE (\%) & \textbf{0.334} & 0.396 & 0.391 & 0.462 & 0.485 \\
    \bottomrule
  \end{tabular}

\end{table}

%% file: Sections/Tables/kd_validation.tex
\begin{table}[htbp]
\centering
\caption{Comparison of mean indicators between the distilled and vanilla models on different protocols}
\label{tab:vanilla_vs_distilled_means_multirow}

\begin{tabular}{lllcccc}
\toprule
Model & Protocol & Regime & MAE & RMSE & R$^2$ & MAPE (\%) \\
\midrule
\multirow{4}{*}{DLinear\cite{zeng2023transformers}}
  & \multirow{2}{*}{CC}   & vanilla   & 0.149 & 0.180 & 0.918 & 1.35 \\
  &                        & distilled & \textbf{0.133} & \textbf{0.162} & \textbf{0.934} & \textbf{1.20} \\
  & \multirow{2}{*}{CCCV} & vanilla   & 0.094 & 0.109 & 0.953 & 0.77 \\
  &                        & distilled & \textbf{0.084} & \textbf{0.097} & \textbf{0.963} & \textbf{0.69} \\
\midrule
\multirow{4}{*}{FiLM\cite{zhou2022film}}
  & \multirow{2}{*}{CC}   & vanilla   & 0.089 & \textbf{0.111} & \textbf{0.968} & 0.80 \\
  &                        & distilled & \textbf{0.063} & 0.114 & 0.963 & \textbf{0.55} \\
  & \multirow{2}{*}{CCCV} & vanilla   & 0.054 & 0.060 & 0.986 & 0.44 \\
  &                        & distilled & \textbf{0.021} & \textbf{0.037} & \textbf{0.994} & \textbf{0.18} \\
\midrule
\multirow{4}{*}{FreTS\cite{yi2023frequency}}
  & \multirow{2}{*}{CC}   & vanilla   & 0.148 & 0.179 & 0.919 & 1.34 \\
  &                        & distilled & \textbf{0.057} & \textbf{0.078} & \textbf{0.983} & \textbf{0.51} \\
  & \multirow{2}{*}{CCCV} & vanilla   & 0.093 & 0.108 & 0.954 & 0.77 \\
  &                        & distilled & \textbf{0.029} & \textbf{0.038} & \textbf{0.994} & \textbf{0.24} \\
\midrule
\multirow{4}{*}{LightTS\cite{zhang2207less}}
  & \multirow{2}{*}{CC}   & vanilla   & \textbf{0.122} & \textbf{0.159} & \textbf{0.935} & \textbf{1.10} \\
  &                        & distilled & 0.132 & 0.167 & 0.928 & 1.20 \\
  & \multirow{2}{*}{CCCV} & vanilla   & \textbf{0.073} & \textbf{0.092} & \textbf{0.966} & \textbf{0.60} \\
  &                        & distilled & 0.083 & 0.100 & 0.961 & 0.68 \\
\midrule
\multirow{4}{*}{PaiFilter\cite{yi2024filternet}}
  & \multirow{2}{*}{CC}   & vanilla   & 0.148 & 0.179 & 0.918 & 1.34 \\
  &                        & distilled & \textbf{0.043} & \textbf{0.067} & \textbf{0.985} & \textbf{0.39} \\
  & \multirow{2}{*}{CCCV} & vanilla   & 0.094 & 0.109 & 0.953 & 0.77 \\
  &                        & distilled & \textbf{0.016} & \textbf{0.027} & \textbf{0.997} & \textbf{0.13} \\
\midrule
\multirow{4}{*}{PatchTST\cite{nie2022time}}
  & \multirow{2}{*}{CC}   & vanilla   & 0.064 & 0.088 & 0.978 & 0.58 \\
  &                        & distilled & \textbf{0.044} & \textbf{0.073} & \textbf{0.981} & \textbf{0.40} \\
  & \multirow{2}{*}{CCCV} & vanilla   & 0.036 & 0.045 & 0.992 & 0.30 \\
  &                        & distilled & \textbf{0.016} & \textbf{0.026} & \textbf{0.997} & \textbf{0.13} \\
\midrule
\multirow{4}{*}{SegRNN\cite{lin2023segrnn}}
  & \multirow{2}{*}{CC}   & vanilla   & 0.080 & 0.108 & 0.970 & 0.72 \\
  &                        & distilled & \textbf{0.065} & \textbf{0.091} & \textbf{0.979} & \textbf{0.59} \\
  & \multirow{2}{*}{CCCV} & vanilla   & 0.047 & 0.058 & 0.987 & 0.39 \\
  &                        & distilled & \textbf{0.038} & \textbf{0.047} & \textbf{0.991} & \textbf{0.31} \\
\midrule
\multirow{4}{*}{TSMixer\cite{chen2023tsmixer}}
  & \multirow{2}{*}{CC}   & vanilla   & 0.148 & 0.181 & 0.917 & 1.33 \\
  &                        & distilled & \textbf{0.113} & \textbf{0.141} & \textbf{0.949} & \textbf{1.02} \\
  & \multirow{2}{*}{CCCV} & vanilla   & 0.092 & 0.108 & 0.954 & 0.76 \\
  &                        & distilled & \textbf{0.071} & \textbf{0.084} & \textbf{0.972} & \textbf{0.58} \\
\bottomrule
\end{tabular}

\end{table}

%% file: Sections/Discussion.tex
\section{Discussion}

\subsection{Distillation Interpretability Analysis}

Knowledge distillation transfers temporal priors from a TSFM into lightweight experts. 
Beyond aggregate error metrics, we aim to understand how distillation reshapes the models’ autoregressive behavior over capacity–degradation histories. 
We therefore conduct an interpretability study to attribute forecast decisions to specific cycles in the lookback context and compare attribution profiles before and after distillation to reveal shifts in temporal dependence.
This analysis comprises two parts: \emph{LIME for time series}, which details our explanation protocol, and \emph{Interpretability results}, which summarizes empirical findings across models and protocols.

\subsubsection{LIME for Time Series}

Let $\{y_t\}_{t=1}^{T}$ be a univariate capacity–degradation series indexed by cycle.
For each window at time $t$, we use a lookback $\mathbf{x}_t=(y_{t-L+1},\ldots,y_t)\in\mathbb{R}^{L}$ with $L=96$ and produce an $H$–step forecast $\hat{\mathbf{y}}_{t+1:t+H}$ with $H=96$ via iterative decoding.

LIME requires a scalar output. 
We adopt the horizon mean as \textbf{Eq.}\ref{eq:forecast_mean}.
\begin{equation}
\label{eq:forecast_mean}
g(\mathbf{x}_t)=\frac{1}{H}\sum_{h=1}^{H}\hat{y}_{t+h},
\end{equation}
which aligns with long-horizon accuracy and stabilizes local fits. 
For any perturbed $\tilde{\mathbf{x}}$, we apply per-instance MinMax scaling, roll out $H$ steps, and compute $g(\tilde{\mathbf{x}})$ using \textbf{Eq.}\ref{eq:forecast_mean}.

Treat $\mathbf{x}_t$ as a tabular instance with $L$ continuous features mapped to $\{t-L+1,\ldots,t\}$.
Around $\mathbf{x}_t$, LIME samples $N=200$ perturbations $\{\mathbf{z}_j\}_{j=1}^{N}$, evaluates $g(\mathbf{z}_j)$, and fits a sparse, locally weighted linear model like \textbf{Eq.}\ref{eq:lime_surrogate}.
\begin{equation}
\label{eq:lime_surrogate}
\tilde{g}_t(\mathbf{z})=\mathbf{w}_t^{\top}\mathbf{z}+b_t,
\end{equation}
by minimizing \textbf{Eq.}\ref{eq:lime_obj}.
\begin{equation}
\label{eq:lime_obj}
\min_{\mathbf{w}_t,b_t}\;
\sum_{j=1}^{N}\pi\!\left(\mathbf{z}_j,\mathbf{x}_t\right)\big(g(\mathbf{z}_j)-\mathbf{w}_t^{\top}\mathbf{z}_j-b_t\big)^2+\Omega(\mathbf{w}_t),
\end{equation}
where $\pi$ is the locality kernel and $\Omega$ regularizes complexity. 
The coefficient $w_{t,k}$ is the local importance of lookback position $k$ (magnitude = strength; sign = direction).
We store $\mathbf{w}_t\in\mathbb{R}^{L}$ for each window to form an attribution matrix $(\#\text{windows})\times L$ and summarize position-wise importance by
\begin{equation}
\label{eq:summary_importance}
\bar{I}_k=\frac{1}{W}\sum_{t=1}^{W}\big|w_{t,k}\big|.
\end{equation}
Heatmaps visualize temporal dependence; window-level MAE, RMSE, and $R^2$ enable linking attribution to accuracy.
The same protocol is applied to vanilla and distilled models for controlled comparison.

\subsubsection{Interpretability result}

\begin{figure}[  htb]
  \centering
  \includegraphics[width=0.75\linewidth]{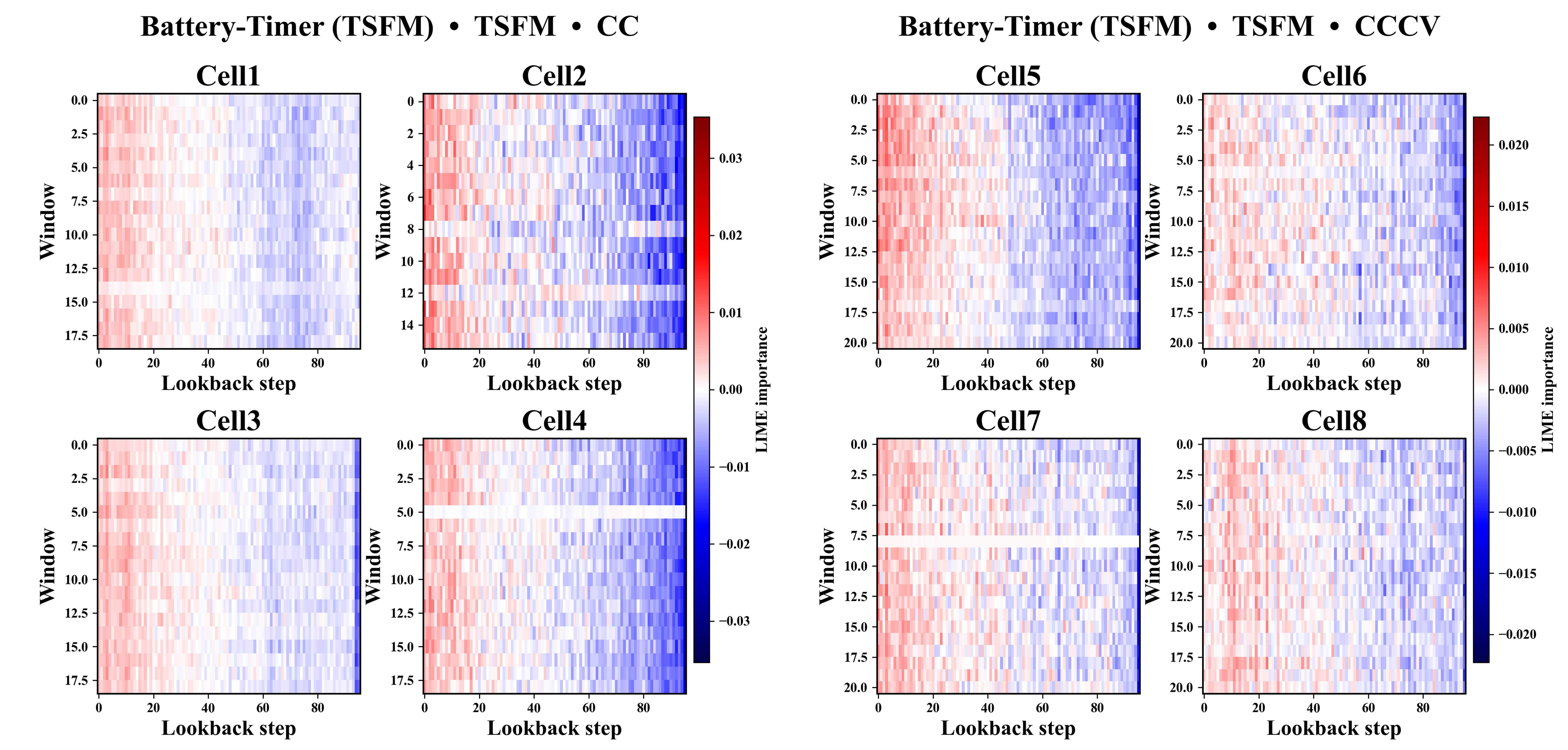}
\caption{LIME attribution heatmaps of Battery-Timer under CC and CCCV protocols. 
Earlier cycles in the lookback window receive predominantly positive attributions, whereas later cycles receive negative attributions, indicating near-term reliance with long-range correction.}
  \label{fig:TSFM_explain}
\end{figure}

Using the LIME protocol described above, we visualize the contribution of historical cycles to Battery-Timer’s capacity-degradation forecasts (Fig.~\ref{fig:TSFM_explain}). 
Under both CC and CCCV protocols, the TSFM assigns predominantly positive attributions to earlier positions in the lookback window and negative attributions to later positions, indicating reliance on recent history with long-range correction. 
Given Battery-Timer’s clear performance advantage over the baselines in Fig.~\ref{c_with_state-of-the-art}, we regard this attribution pattern as a representative and reliable explanation for capacity-degradation forecasting on the CycleLife-SJTUIE dataset.

\begin{figure}[  htb]
  \centering
  \panel[0.45\linewidth]{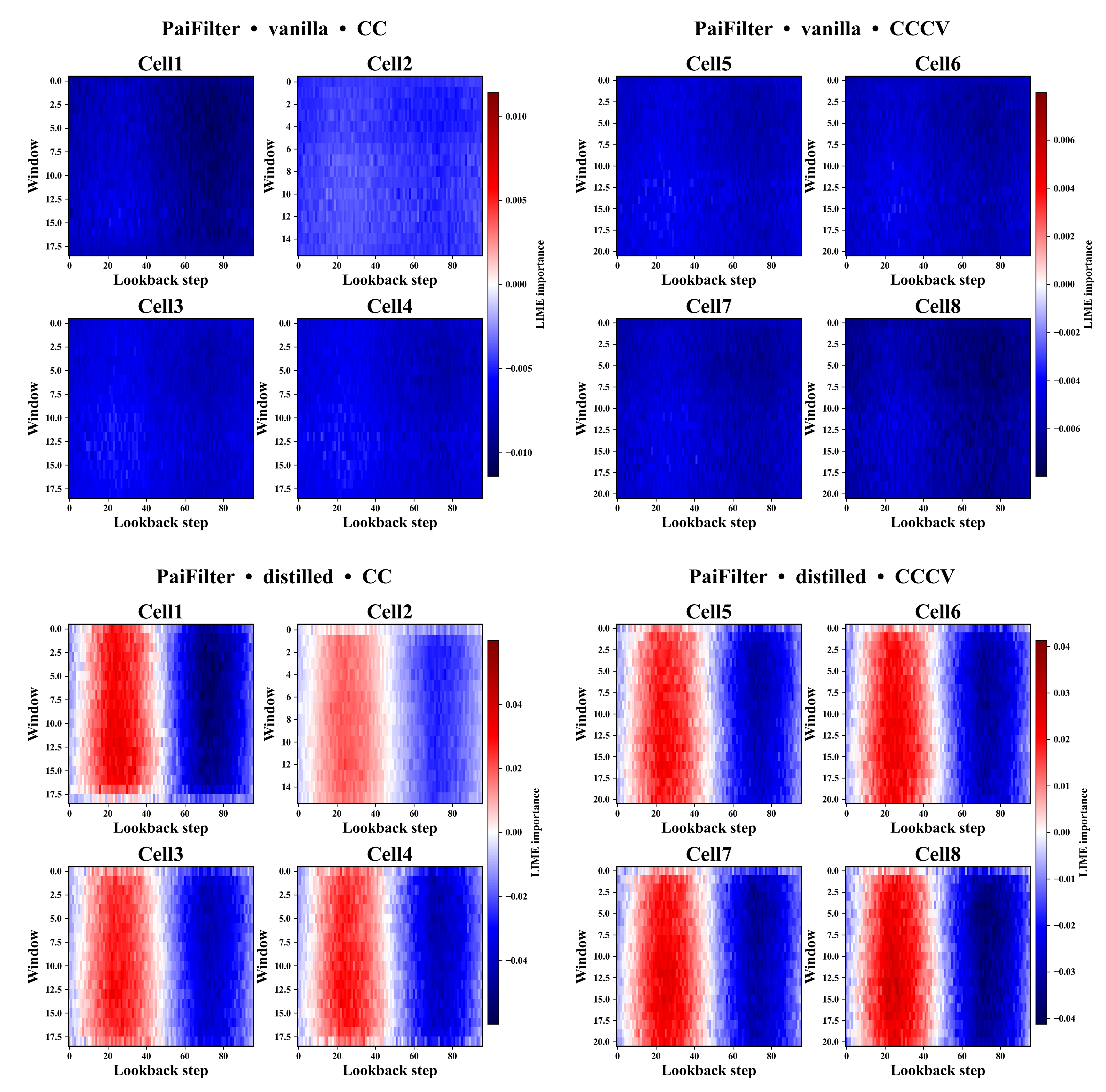}{(a)}
  \panel[0.45\linewidth]{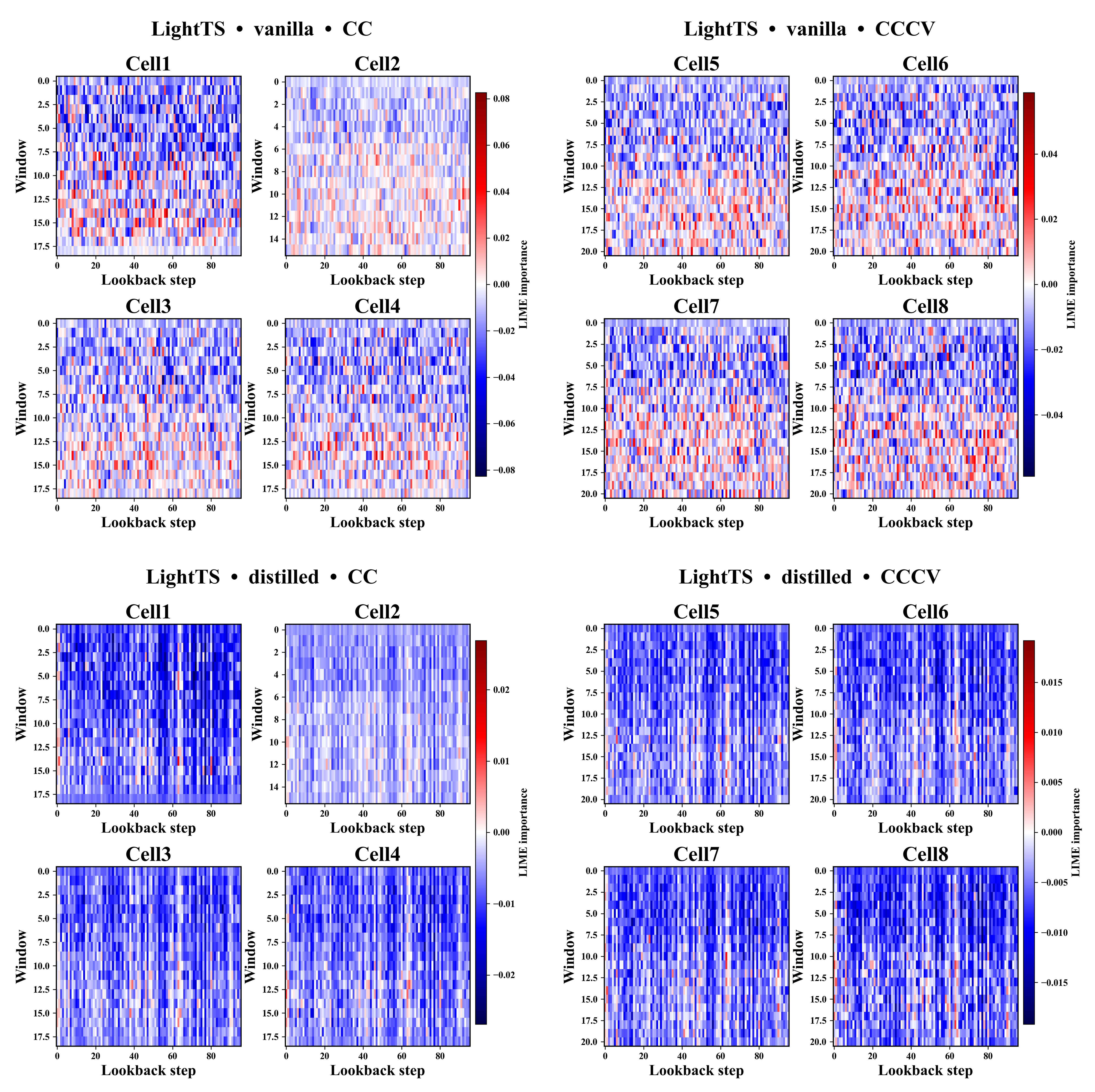}{(b)}
  \caption{\textbf{Representative LIME attribution heatmaps for expert models.} 
  \textbf{(a)} PaiFilter before and after knowledge distillation.
  \textbf{(b)} LightTS before and after knowledge distillation.}
  \label{fig:expert_explain}
\end{figure}

To investigate the interpretability of TSFM knowledge distillation, we compare PaiFilter, which shows the most significant performance improvement after distillation, with LightTS, which experiences a performance decline. 
As shown in Fig.~\ref{fig:expert_explain}, the improvements in PaiFilter due to distillation are clearly reflected in the changes in the LIME heatmap. 
After distillation, the PaiFilter heatmap closely resembles that of Battery-Timer, with positive attributions assigned to earlier cycles and negative attributions to later cycles. 
In contrast, LightTS fails to exhibit effective temporal focus both before and after distillation, which explains its relatively poorer performance in terms of both absolute error values and distillation improvements compared to other models.

This interpretability analysis, by constructing LIME-based temporal attribution maps, highlights how knowledge distillation reshapes the model's reliance on different lags in the capacity degradation sequence.
This contributes to the broader field of battery capacity degradation forecasting by providing explanatory maps of superior model performance, and offers quantitative evidence for evaluating cross-protocol and cross-capacity generalization as well as for designing prior knowledge transfer strategies.

\subsection{Ablation study}

We conduct an ablation study to quantify how design choices in TSFM fine-tuning affect long-horizon capacity forecasting and training efficiency. 
Specifically, we vary (i) the LoRA injection positions within the Transformer block, (ii) the LoRA rank and scaling factor $\alpha$ that jointly control the adaptation capacity and magnitude, and (iii) the trend-penalty coefficient $\lambda$ that encourages monotonic degradation.
Experiments are performed on CycleLife-SJTUIE under CC and CCCV protocols with fixed lookback and horizon ($L=H=96$). 
We introduce the monotonicity violation rate (MVR), defined as the proportion of positive adjacent differences in the forecast sequence $\hat{\mathbf{y}}$ as \textbf{Eq.}\ref{eq:mvr}.

\begin{equation}
\label{eq:mvr}
\mathrm{MVR} \;=\; \frac{1}{H-1}\sum_{t=1}^{H-1}\mathbf{1}\!\left\{\hat{y}_{t+1}-\hat{y}_{t}>0\right\},
\end{equation}

\begin{figure}[  htb]
  \centering
  \includegraphics[width=\linewidth]{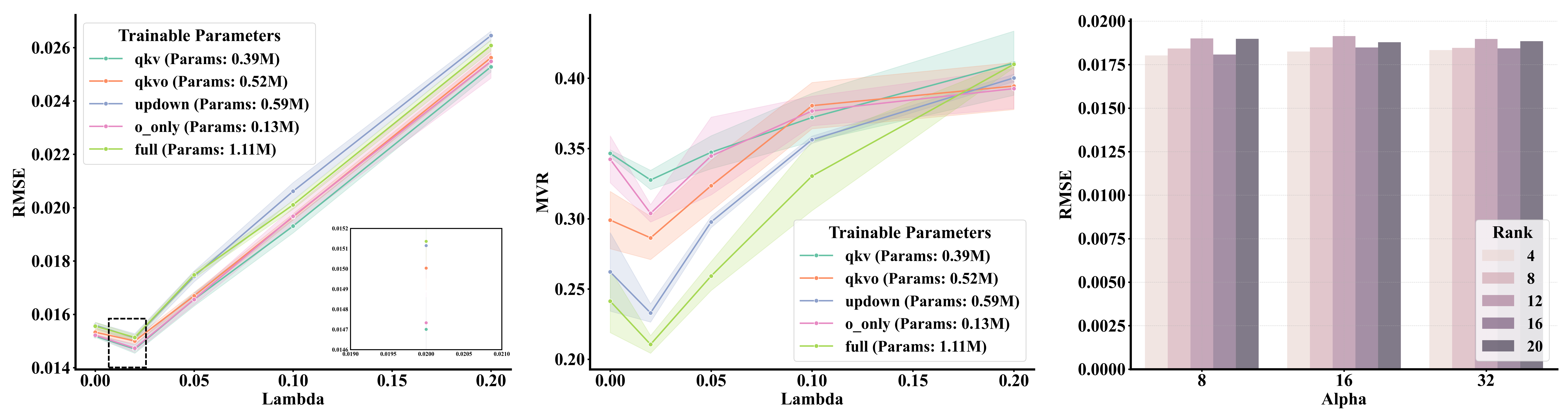}
  \caption{Parameter ablation study: impact of LoRA injection positions, LoRA scaling factor $\alpha$ and rank $r$, and the trend-penalty coefficient $\lambda$.
  All results were obtained from a 3-epoch grid search (per configuration).}
  \label{fig:a_study}
\end{figure}

The results in Fig.~\ref{fig:a_study} reveal configurations that balance accuracy, stability, and parameter efficiency for deployable TSFM fine-tuning.

Ablation results indicate that inserting LoRA into the attention projections ($q, k, v$) yields the best performance. 
The trend-penalty coefficient attains its optimum at $\lambda = 0.02$, a small value consistent with our design goal: the trend term calibrates against upward drift rather than enforcing ever-decreasing slopes. 
For $\lambda > 0.02$, both RMSE and MVR increase, suggesting that overweighting the trend penalty degrades fit and stability. 
Within the tested range, the LoRA scaling factor $\alpha$ and rank $r$ has only a minor effect on accuracy.

\subsection{Failure Modes of Long-Horizon Rolling Forecasts}

\begin{figure}[  htb]
    \centering\renewcommand{\figurename}{Figure}
    \includegraphics[width=0.8\textwidth]{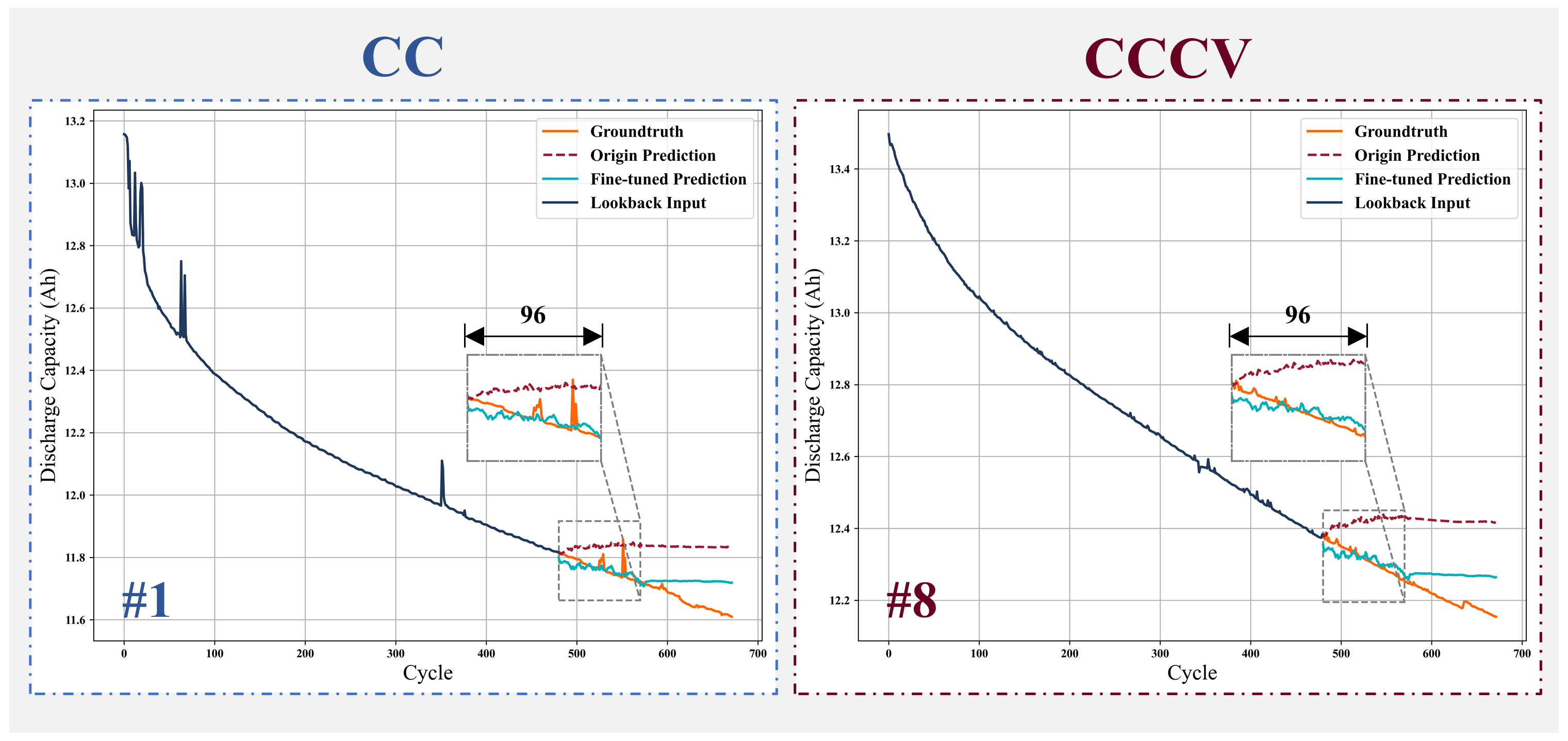}
    \caption{When the output horizon is extended to 192, the predictions beyond step 96 gradually collapse into an over-smoothed, almost linear trajectory and become unreliable for evaluation.}
    \label{img61}
\end{figure}

Existing open-source time-series foundation models have shown that scaling Transformer architectures to temporal data is feasible and effective within a pre-defined output horizon. 
However, our experiments reveal a pronounced failure mode once the forecast horizon extends beyond the pre-trained range. 
As illustrated in Figure~\ref{img61}, when the horizon is increased from 96 to 192 steps, the predictions remain reasonable within the first 96 steps but rapidly deteriorate afterwards. 
The generated sequence becomes progressively over-smoothed, gradually approaches an almost linear or even constant trajectory, and eventually loses practical value for evaluation and decision-making. This degradation persists even after domain-specific fine-tuning, which suggests that it cannot be explained solely by a lack of domain knowledge or insufficient pre-training data.

This behaviour arises from the prevailing training and deployment paradigm of time-series foundation models. 
In typical practice, the model is optimized under a fixed training horizon, for example 96 steps, by minimizing point-wise errors such as mean squared error. 
The model either produces all future steps in a single forward pass, or is unrolled only over a limited number of autoregressive steps during training. 
When a much longer horizon is required at inference time, a natural strategy is to feed the model’s own predictions back as inputs and perform recursive forecasting as \textbf{Eq.}\ref{eq:agg}.

\begin{equation}
\label{eq:agg}
\hat{y}_{t+1} = f(x_{\le t}), \quad
\hat{y}_{t+2} = f(x_{\le t}, \hat{y}_{t+1}), \quad
\hat{y}_{t+3} = f(x_{\le t}, \hat{y}_{t+1}, \hat{y}_{t+2}), \dots
\end{equation}

In this autoregressive mode, the model gradually departs from the training distribution, because the historical context is increasingly dominated by its own prediction errors instead of true observations. This mismatch between training and inference, often described as exposure bias and error accumulation, drives the model into regions of the state space that have not been covered during training. 
Under a squared-error objective and given the intrinsic smoothing tendency of deep sequence models, the resulting behaviour is a contraction toward a stable mean trajectory, so that free-running forecasts converge to a fixed point and empirically collapse into a nearly straight line as the horizon grows.

This failure mode seems difficult to reconcile with the success of large language models, which are also based on autoregressive Transformers yet routinely generate coherent sequences of variable length. Large language models, however, differ from time-series foundation models in several essential aspects. 
They are trained to predict the next token together with a dedicated end-of-sequence (EOS) token on corpora that contain sentences and documents of diverse lengths. 
The stopping rule is thus embedded in the learned conditional distribution, and generation proceeds step by step until the EOS symbol is produced or a prescribed upper bound is reached. 
Moreover, the output space is discrete and highly structured. 
The categorical cross-entropy objective encourages the model to capture rich linguistic regularities rather than collapse to a single average token, and stochastic decoding further mitigates regression toward the mean. Finally, generation is performed purely in the forward direction, without backpropagation through long rollouts at inference time, which avoids the vanishing-gradient effects associated with deeply recursive training.

Motivated by these observations, a natural direction for future work is to endow time-series foundation models with an autoregressive interface analogous to that of large language models and to train them to jointly predict the next time-step value and an explicit termination signal. 
Such a design would allow future trajectories to be generated sequentially with a learned, variable horizon, thereby alleviating the mismatch between training and inference when long-range forecasts are required.

%% file: Sections/Conclusion.tex
\section{Conclusion}

The recent progress in the development of foundation models for time series is encouraging.
However, the application of such models to battery capacity degradation prediction remains limited, especially in the context of domain adaptation for degradation forecasting tasks.
This paper proposed a degradation-aware fine-tuning and distillation framework for lithium-ion battery capacity forecasting built on top of a Timer-based time-series foundation model (TSFM). 
By combining LoRA-based parameter-efficient adaptation, a monotonicity-aware trend penalty, and cross-protocol knowledge distillation into lightweight expert models, we aimed to address two coupled challenges: 
(i) achieving accurate forecasting under heterogeneous datasets and operating protocols, and 
(ii) enabling deployment on resource-constrained battery management systems without retraining from scratch for every new scenario.

On the CycleLife-SJTUIE benchmark, Battery-Timer consistently achieves lower MAE/MAPE and higher $R^2$ than a range of recent time-series forecasters, while also reaching sub-percent MAPE levels that are competitive with, or better than, typical RF/SVM/LSTM baselines reported for similar SOH/RUL tasks. 
This demonstrates that a carefully adapted foundation model can deliver both high accuracy and strong transferability beyond the data it was trained on, providing a more general and reusable degradation prior than task-specific models.

To make this prior deployable, we introduce a TSFM-to-expert knowledge distillation scheme. 
Distilled students retain the compact architectures and inference costs of the original expert models, yet in most cases show substantial reductions in percentage error and gains in $R^2$ on both in-protocol (CCCV) and cross-protocol (CC, zero-shot) evaluations. 
Ablation studies on LoRA injection positions, rank $r$ / $\alpha$, and the trend-penalty coefficient $\lambda_{\text{trend}}$, validate the necessity of the proposed modules.
A LIME-based time-series interpretability analysis further shows that the TSFM and well-distilled experts focus on physically meaningful parts of the degradation history, rather than spurious fluctuations.

Beyond lithium-ion batteries, the proposed paradigm of pretraining a TSFM, applying parameter-efficient, industrial case adaptation, and distilling into compact experts can be transferred to a wide range of long-horizon degradation and survival-type time series. 
For example, in wearable health monitoring, a TSFM trained on large-scale multi-sensor streams (heart rate, activity, respiration) could be distilled into tiny on-device experts for early warning of arrhythmia or heart-failure risk under diverse lifestyles. 
In industrial prognostics, the same idea can be used to learn generic degradation priors from historical vibration and temperature data of bearings, gearboxes, wind turbines, or mining machinery, and then deploy distilled experts tailored to specific assets or operating regimes. 
In energy systems, the framework can be extended to forecast performance decay of photovoltaic modules, fuel-cell stacks, or power-electronic components, where long-term reliability must be monitored at the edge. 
These scenarios share the core challenges targeted in this work, including scarce labeled data in new conditions, distribution shift across fleets, and strict computational budgets, so we expect the Battery-Timer framework and released resources to provide a reusable starting point for foundation-model-based time-series management in many adjacent domains.

%% file: Sections/materials.tex
\section*{Model and Data available}
The Battery-Timer model developed in this study and the Snippet of CycleLife-SJTUIE are available at [\url{https://github.com/sjtu-chan-joey/Battery-Timer}].

%% file: Sections/Acknowledgment.tex
\section*{Acknowledgements}
This work is sponsored by the National Natural Science Foundation of China under Grant 72471143 and 72571178.